\documentclass{article}

\usepackage{hyperref}

\usepackage[accepted]{mlsys2024}

\usepackage{color,xcolor}
\usepackage{epsfig}
\usepackage{graphicx}
\usepackage{subfiles}

\usepackage{adjustbox}
\usepackage{array}
\usepackage{booktabs}
\usepackage{colortbl}
\usepackage{float,wrapfig}
\usepackage{hhline}
\usepackage{multirow}
\usepackage{multirow}
\usepackage{booktabs}
\usepackage{subcaption} %
\usepackage[labelfont={bf},labelsep={period},font={small}]{caption}

\usepackage{tabu}

\let\llncssubparagraph\subparagraph
\let\subparagraph\paragraph
\usepackage[compact]{titlesec}
\let\subparagraph\llncssubparagraph

\usepackage{amsmath,amsfonts,amssymb}
\usepackage{bm}
\usepackage{nicefrac}
\usepackage{microtype}
\usepackage{mathtools}

\usepackage{changepage}
\usepackage{extramarks}
\usepackage{fancyhdr}
\usepackage{lastpage}
\usepackage{setspace}
\usepackage{soul}
\usepackage{xspace}

\usepackage{url}

\usepackage{algpseudocode}
\usepackage{enumerate}

\usepackage{times}
 
\usepackage{pbox}
\usepackage{footnote}
\usepackage{tablefootnote}

\usepackage{paralist}

\usepackage{enumitem}

\definecolor{mydarkgreen}{rgb}{0.02,0.6,0.02}
\usepackage{pifont}

\usepackage[normalem]{ulem}

\usepackage{ dsfont }

\usepackage{pifont}
\usepackage{nccmath}

\usepackage{paralist}
\usepackage{listings}
\usepackage{xcolor}

\definecolor{codegreen}{rgb}{0,0.6,0}
\definecolor{codegray}{rgb}{0.5,0.5,0.5}
\definecolor{codepurple}{rgb}{0.58,0,0.82}
\definecolor{backcolour}{rgb}{0.95,0.95,0.92}

\definecolor{mydarkred}{rgb}{0.8,0.02,0.02}
\definecolor{mydarkorange}{rgb}{0.40,0.2,0.02}
\definecolor{mypurple}{RGB}{111,0,255}
\definecolor{myred}{rgb}{1.0,0.0,0.0}
\definecolor{mygold}{rgb}{0.75,0.6,0.12}
\definecolor{mydarkgray}{rgb}{0.66, 0.66, 0.66}
\definecolor{mygray}{gray}{0.9}

\lstdefinestyle{mystyle}{
    backgroundcolor=\color{backcolour},   
    commentstyle=\color{codegreen},
    keywordstyle=\color{magenta},
    numberstyle=\tiny\color{codegray},
    stringstyle=\color{codepurple},
    basicstyle=\ttfamily\footnotesize,
    breakatwhitespace=false,         
    breaklines=true,                 
    captionpos=b,                    
    keepspaces=true,                 
    numbers=left,                    
    numbersep=5pt,                  
    showspaces=false,                
    showstringspaces=false,
    showtabs=false,                  
    tabsize=2
}
\newcommand{\myparagraph}[1]{\vspace{-5pt}\paragraph{#1}}

\newcommand{\ignorethis}[1]{}

\makeatletter
\DeclareRobustCommand\onedot{\futurelet\@let@token\@onedot}
\def\@onedot{\ifx\@let@token.\else.\null\fi\xspace}

\def\eg{\emph{e.g}\onedot} 
\def\ie{\emph{i.e}\onedot}

\makeatother

\makeatletter

\newcommand\footnoteref[1]{\protected@xdef\@thefnmark{\ref{#1}}\@footnotemark}
\makeatother

\def\method{Activation-aware Weight Quantization\xspace}
\def\methodshort{AWQ\xspace}
\def\system{TinyChat\xspace}

\mlsystitlerunning{AWQ: Activation-aware Weight Quantization for On-Device LLM Compression and Acceleration}

\begin{document}

\twocolumn[
\mlsystitle{AWQ: \underline{A}ctivation-aware \underline{W}eight \underline{Q}uantization for \\On-Device LLM Compression and Acceleration}

\mlsyssetsymbol{equal}{*}
\mlsyssetsymbol{equal-sys}{\dag}

\begin{mlsysauthorlist}
\mlsysauthor{Ji Lin}{equal,mit}
\mlsysauthor{Jiaming Tang}{equal,mit,sjtu}
\mlsysauthor{Haotian Tang}{equal-sys,mit}
\mlsysauthor{Shang Yang}{equal-sys,mit}
\mlsysauthor{Wei-Ming Chen}{nvidia}
\mlsysauthor{Wei-Chen Wang}{mit}
\mlsysauthor{Guangxuan Xiao}{mit}
\mlsysauthor{Xingyu Dang}{mit,tsinghua}
\mlsysauthor{Chuang Gan}{ibm,umass}
\mlsysauthor{Song Han}{mit,nvidia}
\end{mlsysauthorlist}

\mlsysaffiliation{mit}{MIT}
\mlsysaffiliation{sjtu}{Shanghai Jiao Tong University}
\mlsysaffiliation{nvidia}{NVIDIA}
\mlsysaffiliation{tsinghua}{Tsinghua University}
\mlsysaffiliation{ibm}{MIT-IBM Watson AI Lab}
\mlsysaffiliation{umass}{UMass Amherst}

\mlsyscorrespondingauthor{Song Han}{songhan@mit.edu}
\centering
\url{https://github.com/mit-han-lab/llm-awq}

\mlsyskeywords{Machine Learning, MLSys}
\vskip 0.3in
\begin{abstract}

Large language models (LLMs) have transformed numerous AI applications. \textit{On-device} LLM is becoming increasingly important: running LLMs locally on edge devices can reduce the cloud computing cost and protect users' privacy. However, the astronomical model size and the limited hardware resource pose significant deployment challenges. We propose \method (\methodshort), a hardware-friendly approach for LLM low-bit weight-only quantization. \methodshort finds that not all weights in an LLM are equally important. Protecting \textit{only 1\%} salient weights can greatly reduce quantization error. To identify salient weight channels, we should refer to the activation distribution, not weights. 
To avoid the hardware-inefficient mix-precision quantization, we mathematically derive that scaling up the salient channels can reduce the quantization error.
\methodshort employs an equivalent transformation to scale the salient weight channels to protect them. The scale is determined by collecting the activation statistics offline.
\methodshort does not rely on any backpropagation or reconstruction, so it generalizes to different domains and modalities without overfitting the calibration set. \methodshort outperforms existing work on various language modeling and domain-specific benchmarks (coding and math). Thanks to better generalization, it achieves excellent quantization performance for \emph{instruction-tuned} LMs and, for the first time, \emph{multi-modal} LMs. 
Alongside AWQ, we implement \system, an efficient and flexible inference framework tailored for 4-bit on-device LLM/VLMs. With kernel fusion and platform-aware weight packing, \system offers more than \textbf{3}$\times$ speedup over the Huggingface FP16 implementation on both desktop and mobile GPUs. It also democratizes the deployment of the 70B Llama-2 model on mobile GPUs. 

\end{abstract}

]

\newcommand{\mlsysEqualContributionOurs}{\textsuperscript{*}: Algorithm co-lead, \textsuperscript{\dag}: system co-lead.}
\printAffiliationsAndNotice{\mlsysEqualContributionOurs} %
\section{Introduction}

Deploying large language models (LLMs) directly on edge devices is crucial. On-device usage eliminates delays caused by sending data to a cloud server and enables LLMs to operate offline, which is beneficial for real-time applications like virtual assistants, chatbots, and autonomous vehicles. The operational costs associated with maintaining and scaling centralized cloud infrastructure can also be reduced. On-device LLM also enhances data security by keeping sensitive information local, reducing the chance of data breaches. LLMs, grounded in transformer-based architectures~\cite{vaswani2017attention}, have gathered significant attention for their impressive performance across diverse benchmarks~\cite{gpt3, opt, touvron2023llama, scao2022bloom}. However, the large model size leads to the high serving costs. For example, GPT-3 has 175B parameters, which is 350GB in FP16, while the latest B200 GPU only has 192GB memory, let alone edge devices.

\begin{figure}
    \centering
     \includegraphics[width=0.5\textwidth]{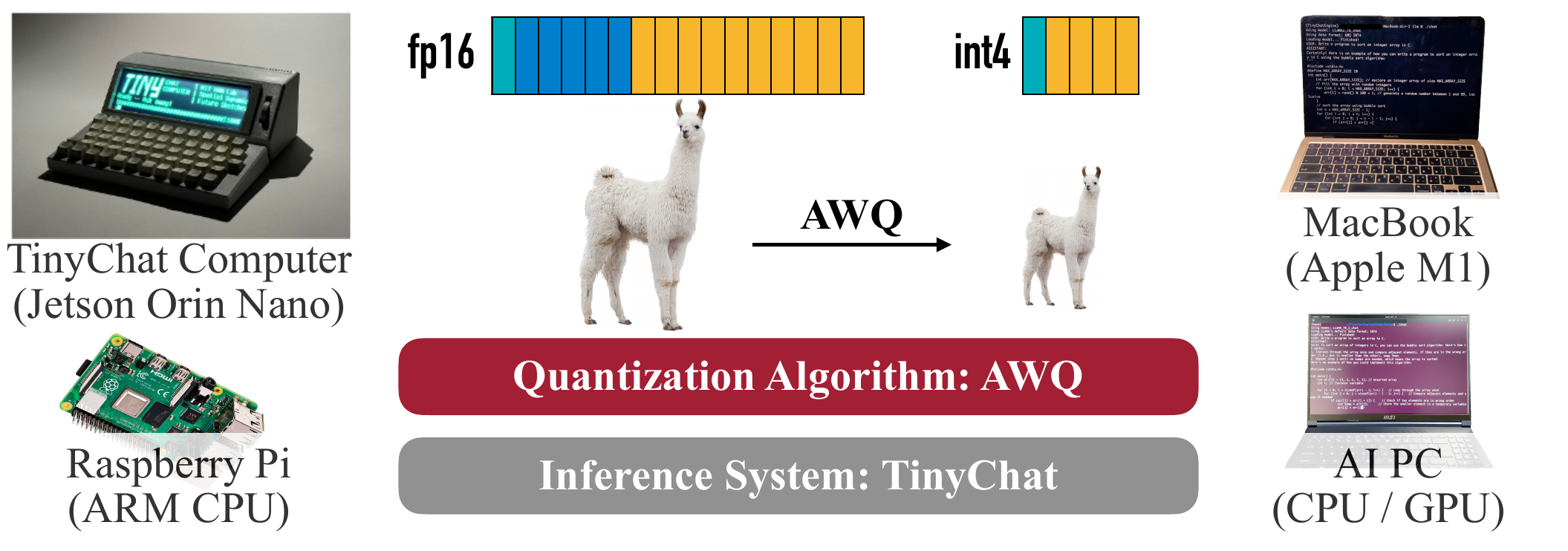}
    \caption{We introduce \textbf{AWQ}, a versatile weight quantization method for LLM. To implement AWQ, we developed \textbf{\system} to deploy 4-bit quantized LLMs into various edge platforms, achieving a \textbf{3-4$\times$} performance boost compared to FP16. Notably, we've also manufactured a \textbf{TinyChat computer}, powered by \system, which contains an NVIDIA Jetson Orin Nano with only 8GB of memory and 15W power consumption. Demo: \url{https://youtu.be/z91a8DrfgEw}.} 
    \label{fig:teaser}
\end{figure}

Low-bit weight quantization for LLMs can significantly reduce the memory footprint of on-device LLM inference but is hard.
Quantization-aware training (QAT) is not efficient due to the high training cost, while post-training quantization (PTQ) suffers from large accuracy degradation under a low-bit setting. The closest work is GPTQ~\cite{frantar2022gptq}, which uses second-order information to perform error compensation. 
However, it may overfit the calibration set during reconstruction, distorting the learned features on out-of-distribution domains (Figure~\ref{fig:calib_set_ablation}), which is problematic since LLMs are \emph{generalist} models. 

In this paper, we propose  \method (\methodshort), a hardware-friendly low-bit weight-only quantization method for LLMs. Our method is based on the observation that \emph{weights are not equally important} for LLMs' performance. There is a small fraction (0.1\%-1\%) of \emph{salient} weights; skipping the quantization of these salient weights will significantly reduce the quantization loss (Table~\ref{tab:fp_ratio}). To find the salient weight channels, the insight is that we should refer to the \emph{activation} distribution instead of the \emph{weight} distribution, despite we are doing \emph{weight-only} quantization: weight channels corresponding to larger activation magnitudes are more salient since they process more important features. 
To avoid the hardware-inefficient mixed-precision implementation, we analyze the error from weight quantization and derive that \emph{scaling up the salient channels can reduce their relative quantization error} (Equation~\ref{eq:scaled_quant}). Following the intuition, we designed a per-channel scaling method to automatically search for the optimal scaling that minimizes the quantization error under full-weight quantization.
\methodshort does not rely on any backpropagation or reconstruction, so it can well preserve LLMs’ generalization ability on various domains and modalities without overfitting to the calibration set. 

To implement AWQ, we designed \system, an efficient inference framework to convert theoretical memory savings from 4-bit LLM to measured speedup. Our framework significantly speeds up linear layers through on-the-fly dequantization. We also take advantage of efficient 4-bit weight packing and kernel fusion to minimize the inference overhead (\eg, intermediate DRAM access and kernel launch overhead), such that we can better realize the speed up from quantizing the weights to 4-bit, despite the computer is byte-aligned.

Experiments show that \methodshort outperforms existing work on various tasks
for  different model families (\eg, LLaMA~\cite{touvron2023llama}, OPT~\cite{opt}) and model sizes.  Thanks to better generalization, it also achieves good quantization performance for \emph{instruction-tuned} LMs (\eg, Vicuna) and, for the first time, \emph{multi-modal} LMs (OpenFlamingo~\cite{openflamingo}). 
\system further translates the $\sim$4$\times$ lower memory footprint to measured speedup. On desktop, laptop and mobile GPUs, we consistently observe a \textbf{3.2-3.3}$\times$ average speedup compared to the FP16 implementation by Huggingface across a diverse spectrum of LLMs. Furthermore, it facilitates effortless deployment of the Llama-2-70B model on a single NVIDIA Jetson Orin with 64GB of memory. It also democratizes 13 billion parameter LLM at an interactive pace of 30 tokens/second on a laptop RTX 4070 GPU with only 8GB of memory.  \methodshort has been widely adopted by industry and open-source community: \href{https://huggingface.co/docs/transformers/main_classes/quantization}{HuggingFace Transformers}, \href{https://github.com/NVIDIA/TensorRT-LLM/}{NVIDIA TensorRT-LLM}, \href{https://blogs.windows.com/windowsdeveloper/2024/05/24/quantization-with-directml-helps-you-scale-further-on-windows/}{Microsfot DirectML}, \href{https://console.cloud.google.com/vertex-ai/publishers/meta/model-garden/llama-2-quantized}{Google Vertex AI}, \href{https://github.com/intel/neural-compressor}{Intel Neural Compressor}, \href{https://aws.amazon.com/blogs/machine-learning/boost-inference-performance-for-llms-with-new-amazon-sagemaker-containers/}{Amazon Sagemaker}, \href{https://community.amd.com/t5/ai/reduce-memory-footprint-and-improve-performance-running-llms-on/ba-p/686157}{AMD}, \href{https://github.com/lm-sys/FastChat/blob/main/docs/awq.md}{FastChat}, \href{https://github.com/vllm-project/vllm/blob/main/vllm/model_executor/layers/quantization/awq.py}{vLLM}, \href{https://github.com/InternLM/lmdeploy}{LMDeploy}, and enables Falcon-180B deployable on a \href{https://github.com/NVIDIA/TensorRT-LLM/blob/main/docs/source/blogs/Falcon180B-H200.md}{single} H200 GPU.

\begin{figure*}[t]
    \centering
     \includegraphics[width=\textwidth]{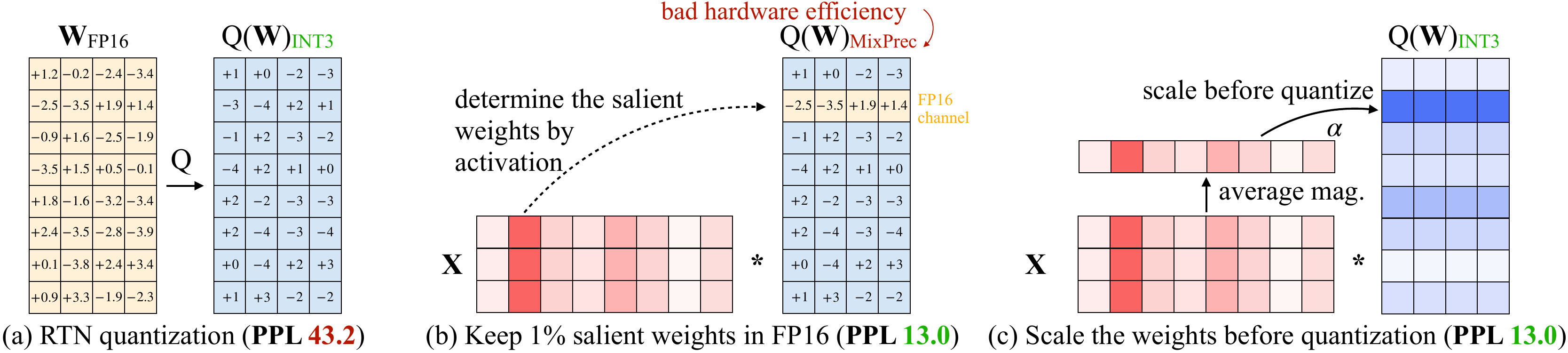}
    \caption{We observe that we can find 1\% of the salient weights in LLMs based on the \emph{activation distribution} (middle). Keeping the salient weights in FP16 can significantly improve the quantized performance (PPL from 43.2 (left) to 13.0 (middle)), but the mixed-precision format is not hardware-efficient. We follow the activation-awareness principle and propose \methodshort (right). \methodshort performs per-channel scaling to protect the salient weights and reduce quantization error. We measure the perplexity of OPT-6.7B under INT3-g128 quantization. %
    } 
    \label{fig:overview}
\end{figure*}

\section{Related Work}
\myparagraph{Model quantization methods. }
Quantization reduces the bit-precision of deep learning models~\cite{han2016deep, jacob2018quantization, nagel2019data, wang2019haq, nagel2020up,lin2020mcunet}, which helps to reduce the model size and accelerate inference. Quantization techniques generally fall into two categories: quantization-aware training (QAT, which relies on backpropagation to update the quantized weights)~\cite{bengio2013estimating, gholami2021survey, nagel2021white, choi2018pact} and post-training quantization~\cite{jacob2018quantization, nagel2019data, nagel2020up} (PTQ, usually training-free). The QAT methods cannot easily scale up to large models like LLMs. Therefore, people usually use PTQ methods to quantize LLMs. 

\myparagraph{Quantization of LLMs.} People study two settings for LLM quantization: (1) W8A8 quantization, where both activation and weights are quantized to INT8~\cite{dettmers2022llmint8, xiao2022smoothquant, zeroquant, outlier_suppression, wei2023outlier}; (2) Low-bit weight-only quantization (\eg, W4A16), where only weights are quantized into low-bit integers~\cite{frantar2022gptq, dettmers2022case, sheng2023high, park2022nuqmm}. We focus on the second setting in this work since it not only reduces the hardware barrier (requiring a smaller memory size) but also speeds up the token generation (remedies memory-bound workload). Apart from the vanilla round-to-nearest baseline (RTN), GPTQ~\cite{frantar2022gptq} is the closest to our work. However, the reconstruction process of GPTQ leads to an over-fitting issue to the calibration set and may not preserve the generalist abilities of LLMs for other modalities and domains. It also requires a reordering trick to work for some models (\eg, LLaMA-7B~\cite{touvron2023llama} and OPT-66B~\cite{opt}). Apart from quantiztion methods designed for general-purporse hardware, SpAtten~\cite{spatten} designs a progressive approach to gradually increase the number of bits used in softmax calculation.

\myparagraph{System support for low-bit quantized LLMs. } Low-bit quantized LLMs have been a popular setting to reduce inference costs. There are some system supports to achieve a practical speed-up. GPTQ~\cite{frantar2022gptq} provides INT3 kernels for OPT models and \texttt{GPTQ-for-LLaMA} extends kernel support for INT4 reordered quantization with the help of Triton~\cite{tillet2019triton}. FlexGen~\cite{sheng2023high}, \texttt{llama.cpp}\footnote{https://github.com/ggerganov/llama.cpp} and \texttt{exllama}\footnote{https://github.com/turboderp/exllama} perform group-wise INT4 quantization to reduce I/O costs and offloading. FasterTransformer implements FP16$\times$INT4 GEMM for weight-only per-tensor quantization but does not support group quantization. LUT-GEMM~\cite{park2022nuqmm} performs bitwise computation on GPU CUDA cores with the help of lookup tables. Our concurrent work, MLC-LLM~\cite{mlc-llm} offers strong results on multiple edge CPU and GPU platforms thanks to the powerful TVM~\cite{chen2018tvm,feng2022tensorir} backend. %

\section{\methodshort: \method}

\emph{Quantization} maps a floating-point number into lower-bit integers. It is an effective method to reduce the model size and inference costs of LLMs~\cite{dettmers2022llmint8, frantar2022gptq, zeroquant, xiao2022smoothquant}. In this section, we first propose a weight-only quantization method to improve accuracy \emph{without training/regression} by protecting more ``important'' weights. And then develop a data-driven method to search for the optimal scaling that reduces quantization errors (Figure~\ref{fig:overview}). 

\subsection{Improving LLM Quantization by Preserving 1\% Salient Weights}
\renewcommand \arraystretch{1.05}
\begin{table*}[t]
    \setlength{\tabcolsep}{4.5pt}
    \small
    \centering
    \vspace{5pt}
    \begin{tabular}{lccccccccccc}
        \toprule
          \multirow{2}{*}{\textbf{PPL $\downarrow$}}   & \multirow{2}{*}{FP16} & \multirow{2}{*}{RTN}   & \multicolumn{3}{c}{FP16\% (based on act.)}  & \multicolumn{3}{c}{FP16\% (based on W)} &  \multicolumn{3}{c}{FP16\% (random)} \\ \cmidrule(lr){4-6} \cmidrule(lr){7-9}  \cmidrule(lr){10-12}
         & & (w3-g128) & 0.1\% & 1\% & 3\% &   0.1\% & 1\% & 3\% &  0.1\% & 1\% & 3\%   \\
        \midrule
        OPT-1.3B  & 14.62 & 119.00 & 25.03 & \textcolor{codegreen}{16.91} & \textcolor{codegreen}{16.68} & 108.71 & 98.55 & 98.08 & 119.76 & 109.38 & 61.49 \\
        OPT-6.7B & 10.86 & 23.54 & \textcolor{codegreen}{11.58} & \textcolor{codegreen}{11.39} & \textcolor{codegreen}{11.36} & 23.41 & 22.37 & 22.45 & 23.54 & 24.23 & 24.22 \\
        OPT-13B &  10.13 & 46.04 &\textcolor{codegreen}{10.51} & \textcolor{codegreen}{10.43} & \textcolor{codegreen}{10.42} & 46.07 & 48.96 & 54.49 & 44.87 & 42.00 & 39.71 \\
        \bottomrule
    \end{tabular}
    \caption{Keeping a small fraction of weights (0.1\%-1\%) in FP16 significantly improves the performance of the quantized models over round-to-nearest (RTN). It is only effective when we select the important weights in FP16 by looking at \emph{activation} distribution instead of \emph{weight} distribution. We highlight results with a decent perplexity in \textcolor{codegreen}{green}.  
    We used INT3 quantization with a group size of 128 and measured the WikiText perplexity ($\downarrow$).  
    } 
    \label{tab:fp_ratio}
\end{table*}

We observe that the weights of LLMs are \emph{not equally important}: there is a small fraction of \emph{salient} weights that are much more important for LLMs' performance compared to others. Skipping the quantization of these salient weights can help bridge the performance degradation due to the quantization loss \emph{without} any training or regression (Figure~\ref{fig:overview}(b)).
To verify the idea, we benchmark the performance of quantized LLMs when skipping part of the weight channels in Table~\ref{tab:fp_ratio}. We measured the performance of INT3 quantized models while keeping some ratios of weight channels in FP16. A widely used method to determine the importance of weights is to look at its magnitude or $L_2$-norm~\cite{han2015learning, frankle2018lottery}. But we find skipping the weight channels with large norm (\ie, FP16\% (based on W)) does not significantly improve the quantized performance, leading to a similar marginal improvement as random selection. 
Interestingly, selecting weights based on \emph{activation magnitude} can significantly improve the performance despite keeping only 0.1\%-1\% of channels in FP16. 
We hypothesize that the input features with larger magnitudes are generally more important. Keeping the corresponding weights in FP16 can preserve those features, which contributes to better model performance. 

\textbf{Limitations: }
Despite keeping 0.1\% of weights in FP16 can improve the quantized performance without a noticeable increase in model size (measured in total bits), such a mixed-precision data type will make the system implementation difficult. We need to come up with a method to protect the important weights without actually keeping them as FP16.

\subsection{Protecting Salient Weights by Activation-aware Scaling}
We propose an alternative method to reduce the quantization error of the salient weight by \emph{per-channel scaling}, which does not suffer from the hardware inefficiency issue. 

\textbf{Analyzing the quantization error.}

\begin{table}
    \setlength{\tabcolsep}{2pt}
    \small
    \centering
    \begin{tabular}{lccccc}
        \toprule
     \textbf{OPT-6.7B}   & $s=1$ & $s=1.25$ & $s=1.5$ & $s=2$ & $s=4$ \\ \midrule
    proportion of $\Delta^{'} \neq \Delta$ & 0\% & 2.8\% & 4.4\% & 8.2\% & 21.2\% \\
    average $\Delta^{'}/ \Delta$ & 1 & 1.005 & 1.013 & 1.038 & 1.213 \\
    average $\frac{\Delta^{'}}{\Delta} \cdot \frac{1}{s}$ & 1 & 0.804  & 0.676 & 0.519 & \textbf{0.303} \\ \midrule 
    Wiki-2 PPL & 23.54 & 12.87 & 12.48 & \textbf{11.92} & 12.36 \\
    \bottomrule
    \end{tabular}
    \caption{Statistics when multiplying the 1\% salient channels by $s>1$. Scaling up the salient channels significantly improves the perplexity (23.54 to 11.92). As $s$ goes larger, the percentage of changed $\Delta$ increases, and the error reduction rate for salient channels also increases. However, the best perplexity is achieved at $s=2$, since further increasing $s$ will increase the quantization error for \emph{non-salient} channels. 
    } 
    \label{tab:scale_study}
\end{table}

We start by analyzing the error from weight-only quantization. 
Consider a group/block of weight $\mathbf{w}$; the linear operation can be written as $y=\mathbf{w}\mathbf{x}$, and the quantized counterpart is $y=Q(\mathbf{w})\mathbf{x}$. Specifically, the quantization function is defined as:
\begin{equation}
    Q(\mathbf{w}) = \Delta \cdot \text{Round}(\frac{\mathbf{w}}{\Delta}), \quad \Delta = \frac{\max(|\mathbf{w}|)}{2^{N-1}},
    \label{eq:org_quant}
\end{equation}
where $N$ is the number of quantization bits, and $\Delta$ is the quantization scaler determined by the absolute maximum value. Now consider a weight element $w\in \mathbf{w}$, if we multiply $w$ with $s> 1$ and the inversely scale $x$, we will have $Q(w\cdot s)(x/s)$, which is:
\begin{equation}
    Q(w\cdot s)\cdot \frac{x}{s} = \Delta^{'} \cdot \text{Round}(\frac{ws}{\Delta^{'}}) \cdot x \cdot \frac{1}{s},
    \label{eq:scaled_quant}
\end{equation}
where $\Delta^{'}$ is the new quantization scaler after applying $s$. We empirically find that: (1) The expected error from $\text{Round}(\cdot)$ (denoted as $\texttt{RoundErr}(\cdot)$) does not change: since the round function maps a floating-point number to an integer, the error is roughly uniformly distributed from [0,0.5], resulting in an average error of $0.25$; i.e., $\texttt{RoundErr}(\cdot)\sim 0.25$. (2) Scaling up a single element $w$ usually does not change the maximum value from the group $\mathbf{w}$. Therefore we have $\Delta^{'}\approx\Delta$;
(3) As $\Delta$ and $x$ are represented in FP16, they have no quantization error. Consequently, the quantization error from equation~\ref{eq:org_quant} and ~\ref{eq:scaled_quant} can be expressed as 
\begin{equation}
\begin{aligned}
\texttt{Err}(Q(w) x) &= \Delta\cdot \texttt{RoundErr}(\frac{w}{\Delta})\cdot x \\
\texttt{Err}(Q(w \cdot s)(\frac{x}{s})) &= \Delta^{'}\cdot \texttt{RoundErr}(\frac{ws}{\Delta^{'}})\cdot x\cdot \frac{1}{s}
\end{aligned}
\end{equation}
The ratio of the new error to the original error is $\frac{\Delta^{'}}{\Delta} \cdot \frac{1}{s}$. Given $\Delta^{'}\approx\Delta$ and $s>1$, the relative error is smaller for the salient weight $w$.

To verify the idea, we multiply the 1\% salient channels with $s>1$ for the OPT-6.7B model, and measure the change in $\Delta$ for each group in Table \ref{tab:scale_study}. We find that scaling up the salient channels is quite effective: the perplexity improves from 23.54 for $s=1$ (simply RTN) to 11.92 for $s=2$. 
As $s$ goes larger, the percentage of changed $\Delta$ generally gets larger, but the percentage is still quite small for $s<2$ (less than 5\%); the relative error for the salient channels continues to go smaller as $s$ increases. Nonetheless, the best PPL actually appears at $s=2$. This is because if we use a very large $s$, it will increase the relative error for the \emph{non-salient} channels when $\Delta$ increases (the error of non-salient channels will be amplified by $\frac{\Delta^{'}}{\Delta}$, and the ratio is larger than 1 for 21.2\% of the channels under $s=4$), which can damage the model's overall accuracy. Therefore, we need to also consider the error from non-salient channels when protecting salient ones.

\begin{table}
    \setlength{\tabcolsep}{5pt}
    \small
    \centering
    \begin{tabular}{llccccccc}
      \toprule
        \textbf{OPT (PPL}$\downarrow$) & 1.3B & 2.7B & 6.7B  & 13B & 30B   \\  \midrule
      FP16 & 14.62 & 12.47 & 10.86 & 10.13	& 9.56 \\ \midrule
      RTN & 119.47 & 298.00 & 23.54 & 46.04	& 18.80	 \\
       1\% FP16 & 16.91 & 13.69 & \textbf{11.39} & \textbf{10.43} & 9.85 \\
       $s=2$ & 18.63 & 14.94 & 11.92 & 10.80 & 10.32 \\
       \methodshort  &  \textbf{16.32} & \textbf{13.58} &\textbf{11.39} & 10.56 & \textbf{9.77}	\\
      \bottomrule
    \end{tabular}
    \caption{\methodshort protects salient weights and reduces quantization error by using a scaling-based method. It consistently outperforms Round-to-nearest quantization (RTN) and achieves comparable performance as mixed-precision (1\% FP16) while being more hardware-friendly. We use 3-bit quantization with group size 128.}
    \label{tab:opt_ppl}
\end{table}

\begin{figure*}[t]
    \centering
     \includegraphics[width=\textwidth]{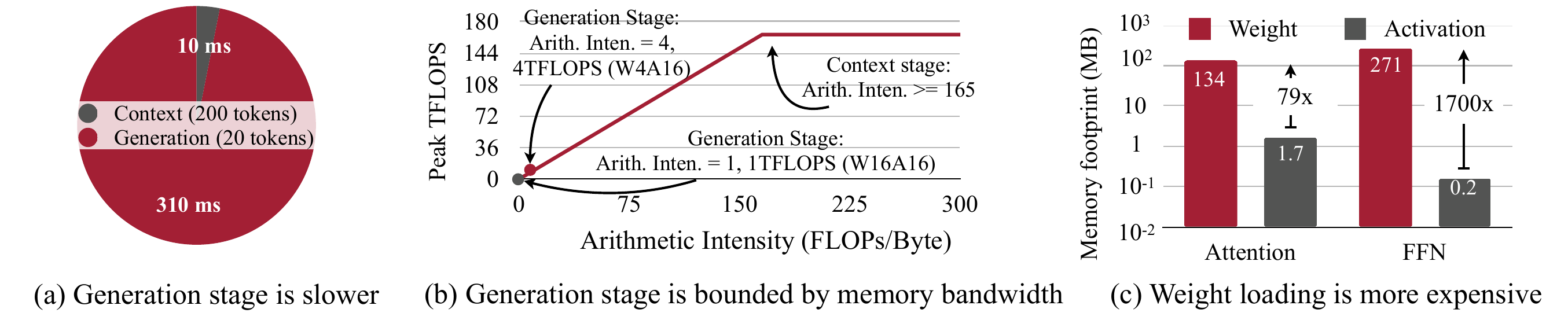}
    \caption{Bottleneck analysis for Llama-2-7B on NVIDIA RTX 4090. \textbf{Left}: In on-device LLM applications, generation stage is much slower than the context stage. \textbf{Middle}: The generation stage is memory bound and has low arithmetic intensity. W4A16 quantization can effectively improve the arithmetic intensity by 4$\times$. \textbf{Right}: The amount of weight access is orders of magnitude larger than the amount of activation access. Thus, weight-only quantization is more effective for on-device LLMs.} 
    \label{fig:memory_bound_analysis}
\end{figure*}

\textbf{Searching to scale. }
To consider both salient and non-salient weights, we choose to automatically search for an optimal (per input channel) scaling factor that minimizes the output difference after quantization for a certain layer. Formally, we want to optimize the following objective:
\begin{equation}
\begin{aligned}
    \mathbf{s}^* &= \mathop{\arg\min}_{\mathbf{s}} \mathcal{L}(\mathbf{s}) \\  
    \mathcal{L}(\mathbf{s}) = \lVert Q(\mathbf{W}\cdot \text{di}&\text{ag}(\mathbf{s}))  (\mathbf{\text{diag}(s)^{-1}} \cdot \mathbf{X}) - \mathbf{W}\mathbf{X}  \rVert
\end{aligned}
\end{equation}
Here $Q$ means the weight quantization function (\eg, INT3/INT4 quantization with group size 128), $\mathbf{W}$ is the original weights in FP16, and $\mathbf{X}$ is the input features cached from a small calibration set (we take a small calibration set from he pre-training dataset in order not to overfit to a specific task). $\mathbf{s}$ is a per-(input) channel scaling factor; for $\mathbf{s^{-1}} \cdot \mathbf{X}$, it can usually be fused into the previous operator~\cite{wei2022outlier, xiao2022smoothquant}.
Since the quantization function is not differentiable, we are not able to directly optimize the problem with vanilla backpropagation. There are some techniques relying on approximated gradients~\cite{bengio2013estimating, esser2019learned}, which we found still suffers from unstable convergence.  

To make the process more stable, we define a \emph{search space} for the optimal scale by analyzing the factors that will affect the choice of scaling factor. As shown in the last section, the saliency of weight channels is actually determined by the activation scale (thus ``activation-awareness''). Therefore, we simply use a very simple search space:
\begin{equation}
\label{eq:scale_search_formula}
    \mathbf{s}=\mathbf{s_X}^{\alpha}, \quad \alpha^*=\mathop{\arg\min}_{\alpha}\mathcal{L}(\mathbf{s_X}^{\alpha})
\end{equation}
$\mathbf{s_X}$ is the average magnitude of activation (per-channel), and we use a single hyper-parameter $\alpha$ to balance between the protection of salient and non-salient channels. We can find the best $\alpha$ by a fast grid search over the interval of $[0, 1]$ ($0$ means we do not scale; $1$  corresponds to the most aggressive scaling in our search space). We further apply weight clipping to minimize the MSE error of quantization. We provide an ablation study on OPT models under INT3-g128 quantization in Table~\ref{tab:opt_ppl}; \methodshort consistently outperforms round-to-nearest quantization (RTN) and achieves comparable performance as mixed-precision (1\% FP16) while being more hardware-friendly.

\textbf{Advantages.} 
Our method does not rely on any regression~\cite{frantar2022gptq} or backpropagation, which is required by many quantization-aware training methods. It has minimal reliance on the calibration set since we only measure the average magnitude per channel, thus preventing over-fitting (Figure~\ref{fig:calib_set_ablation}). Therefore, our method requires fewer data for the quantization process and can preserve LLMs' knowledge outside of the calibration set's distribution. 
See Section~\ref{sec:ablation_study} for more details. %

\section{\system: Mapping AWQ onto Edge Platforms}

AWQ can substantially reduce the size of LLMs. However, converting the theoretical memory savings from W4A16 (4-bit weight, 16-bit activation) quantization into measured speedup is non-trivial. Alternative W8A8 quantization methods, such as SmoothQuant~\cite{xiao2022smoothquant}, maintain \textit{the same} data precision for both storage and computation. This allows the dequantization procedure to be seamlessly integrated into the computation kernel's epilogue. On the other hand, W4A16 quantization employs \textit{different} data types for memory access and computation. As a result, its dequantization must be incorporated into the primary computation loop for optimal performance, posing implementation challenges. To tackle this, we introduce \system: a nimble system for AWQ model inference. It boasts a PyTorch frontend and a backend harnessing device-specific instruction sets (e.g., CUDA/PTX, Neon, AVX).

\subsection{Why AWQ Helps Accelerate On-Device LLMs}

\begin{figure*}
    \centering
     \includegraphics[width=\textwidth]{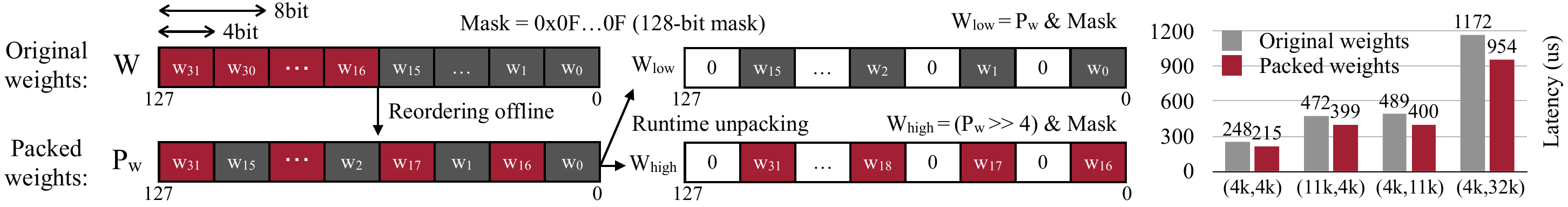}
    \caption{SIMD-aware weight packing for ARM NEON with 128-bit SIMD units. Original weights are reordered and packed to align with the bit width so that the weights can be unpacked into bytes at runtime using AND and shift bitwise operations with a 128-bit mask.} %
    \label{fig:weight_packing}
\end{figure*}

To understand the acceleration opportunities in quantized LLMs on the edge, we start by profiling the latency breakdown of LLaMA-7B~\cite{touvron2023llama} model on an RTX 4090 GPU. We adopt an inference batch size of 1, catering for edge use cases, and implement the model in FP16 with NVIDIA FasterTransformer. 

\myparagraph{Context \textit{vs} generation latency.} As in Figure~\ref{fig:memory_bound_analysis}(a), it takes 310 ms to generate 20 tokens, while summarizing a prompt with 200 tokens only takes 10 ms. Consequently, the generation phase is substantially slower than the context stage, particularly for on-device interactive applications. 

\myparagraph{Generation stage is memory-bound.} To accelerate the generation phase, we conduct a roofline analysis in Figure~\ref{fig:memory_bound_analysis}(b). The 4090 GPU has a peak computation throughput of 165 TFLOPS and a memory bandwidth of 1TB/s. Therefore, any workload with arithmetic intensity (the ratio of FLOPs to memory access) less than 165 is  memory bounded on 4090 GPUs. Notably, when executed in FP16, the generation stage for on-device LLMs has arithmetic intensity$\approx$1. This underscores the memory-bound nature of the workload. Since the FLOPs of a given model is fixed, the only way to improve the peak performance is to reduce the total amount of memory traffic. AWQ reduces the weight memory by four times.

\myparagraph{Weight access dominates memory traffic.}
 We therefore further break down the memory access for weight and activation in Figure~\ref{fig:memory_bound_analysis}(c). Clearly, weight access dominates the memory traffic for on-device LLMs. Quantizing the model weights to 4 bit integers will approximately increase the arithmetic intensity to 4 FLOPs/Byte, leading to a 4TFLOPS peak performance in Figure~\ref{fig:memory_bound_analysis}(b). Since weight-only quantization leads to a lower bit width for weights (and thus higher theoretical performance upper bound), it is natural for AWQ to follow this setting for on-device LLM applications.

\subsection{Deploy AWQ with \system}

To this end, we demonstrated that 4-bit weight quantization could lead to a 4$\times$ theoretical peak performance. We further design \system to realize this speedup. On GPUs, we only focus on implementing essential components, including attention, layer normalization, and linear projection kernels. The flexible frontend allows easy customization and fast support for new models. \system with 4-bit AWQ achieves more than \textbf{3$\times$} speedup compared with the Huggingface FP16 implementation across different families of LLMs on GPUs. On CPUs, we lower the entire computation graph to C++ to minimize overhead.

\myparagraph{On-the-fly weight dequantization.} For quantized layers, as the hardware does not provide multiplication instructions between INT4 and FP16, we need to dequantize the integers to FP16 before performing matrix computation. We avoid writing dequantized weights into DRAM by fusing dequantization kernels with the matrix multplication kernel. Note that such fusion is adopted for both matrix-matrix (MM) and matrix-vector (MV) product kernels. %

\myparagraph{SIMD-aware weight packing.} On-the-fly weight dequantization reduces intermediate DRAM access, but remains expensive. For instance, dequantizing \textit{a single 4-bit weight} involves 1 shift, 1 bitwise AND, and 1 FMA scaling operations, while the dequantized weight undergoes only 1 FMA computation. This process is particularly costly on CPUs with SIMD architecture that favor vectorized instructions. To mitigate this, we suggest platform-specific weight packing tailored to the bitwidth of a device's SIMD units. Figure~\ref{fig:weight_packing} demonstrates our strategy for ARM CPUs with 128-bit SIMD registers offering up to 1.2$\times$ speedup. Here, each register holds 32 4-bit weights, sequenced as $w_0, w_{16}, w_{1}, w_{17}, ..., w_{15}, w_{31}$. This approach requires just three SIMD instructions to unpack \textit{all 32 weights}, as opposed to 3 scalar instructions \textit{per weight} in a conventional packing ($w_0, w_1, ..., w_{31}$). Generally, for $2^n$-bit SIMD registers, adjacent weights will have indices off by $1/8\times2^n$, since each register can hold $1/8\times2^n$ 8-bit integers. On GPUs, we found it more efficient to pack each 8 weights into $w_{\{0,2,4,6,1,3,5,7\}}$ following~\cite{kim2022says}. 

\myparagraph{Kernel fusion.} We also extensively apply kernel fusion to optimize on-device LLM inference. For layer normalization, we fuse all operators (\eg multiplication, division and square root) into a single kernel. For attention layers, we fuse QKV projections into a single kernel, and also perform on-the-fly positional embedding calculation. We also pre-allocate KV caches and perform cache updates within the attention kernel. Kernel fusion is particularly useful for models with inefficient forward pass implementations, such as Falcon~\cite{penedo2023refinedweb} and StarCoder~\cite{li2023starcoder}. Notably, the computation time for each FP16 kernel is in the order of 0.01ms on the 4090 GPU, comparable to the GPU kernel launch overhead. Hence, reducing number of kernel calls through kernel fusion leads to direct speedups.

\section{Experiments}
\begin{table*}
    \setlength{\tabcolsep}{12pt}
    \small
    \centering
    \begin{tabular}{llccccccc}
        \toprule
          \multirow{2}{*}{\textbf{PPL}$\downarrow$} & & \multicolumn{3}{c}{\textbf{Llama-2}} & \multicolumn{4}{c}{\textbf{LLaMA}}  \\  \cmidrule(lr){3-5} \cmidrule(lr){6-9}
          & & 7B & 13B & 70B & 7B  & 13B & 30B & 65B\\ \midrule
        FP16 & - & 5.47 & 4.88 & 3.32 & 5.68 &	5.09 & 4.10	& 3.53\\ \midrule
        \multirow{4}{*}{\shortstack{INT3\\g128}} & RTN & 6.66 &	5.52 & 3.98 & 7.01 &	5.88 & 4.88	& 4.24  \\
        & GPTQ &  6.43	& 5.48 & 3.88 & 8.81 & 5.66 & 4.88 & 4.17\\
         & GPTQ-R &  6.42	& 5.41 & 3.86 & 6.53 & 5.64 & 4.74 & 4.21\\
         & \methodshort  & \textbf{6.24} & \textbf{5.32} & \textbf{3.74} & \textbf{6.35} & \textbf{5.52} & \textbf{4.61} & \textbf{3.95} \\ \midrule
         \multirow{4}{*}{\shortstack{INT4\\g128}} & RTN & 5.73 & 4.98 & 3.46& 5.96 & 5.25 & 4.23 & 3.67\\
         & GPTQ &  5.69	& 4.98 & 3.42 & 6.22 & 5.23 & 4.24 & 3.66\\
         & GPTQ-R & 5.63 & 4.99 & 3.43 & 5.83 & 5.20 & 4.22 & 3.66\\
         & \methodshort & \textbf{5.60} & \textbf{4.97} & \textbf{3.41}  & \textbf{5.78} & \textbf{5.19} & \textbf{4.21} & \textbf{3.62}\\
        \bottomrule
    \end{tabular}
    \caption{\methodshort improves over round-to-nearest quantization (RTN) for different model sizes and different bit-precisions. It consistently achieves better perplexity than GPTQ (w/ and w/o reordering) on LLaMA \& Llama-2 models.
    }
    \label{tab:llama_opt_ppl}
\end{table*}

\begin{table}
    \setlength{\tabcolsep}{5pt}
    \small
    \centering
    \begin{tabular}{lcccccccc}
      \toprule
        \textbf{Wikitext2 PPL}$\downarrow$ & Mixtral-8x7B  & Mistral-7B \\  \midrule
      FP16 & 5.94 & 4.14 \\ \midrule
      INT4-g128 & 6.05 & 4.30	 \\
       INT3-g128 & 6.52 & 4.83  \\
      \bottomrule
    \end{tabular}
    \caption{\methodshort quantization results on Mistral-7B-Instruct-v0.2\cite{jiang2023mistral} and Mixtral-8x7B-Instruct-v0.1 model ~\cite{jiang2024mixtral}. The PPL result on wikitext shows that \methodshort can achieve superior quantization performance on different model architectures including LLMs with GQA and Mixture-of-Experts (MoE) models.}
    \label{tab:opt_ppl}
\end{table}

\subsection{Settings}

\myparagraph{Quantization. } We focus on \emph{weight-only grouped} quantization in this work. As shown in previous work~\cite{dettmers2022case, frantar2022gptq}, grouped quantization is always helpful for improving performance/model size trade-off. We used a group size of 128 throughout the work, except otherwise specified. We focus on INT4/INT3 quantization since they are able to mostly preserve the LLMs' performance~\cite{dettmers2022case}. For \methodshort, we used a small calibration set from the Pile~\cite{gao2020pile} dataset in order not to overfit to a specific downstream domain. We used a grid size of 20 to search for the optimal $\alpha$ in Equation~\ref{eq:scale_search_formula}. 

\myparagraph{Models. } We benchmarked our method on LLaMA~\cite{touvron2023llama} and OPT~\cite{opt} families. There are other open LLMs like BLOOM~\cite{scao2022bloom}, but they are generally worse in quality, so we do not include them in our study. We further benchmark an instruction-tuned model Vicuna~\cite{vicuna2023} and visual language models OpenFlamingo-9B~\cite{openflamingo} and LLaVA-13B~\cite{liu2023llava} to demonstrate the generability of our method. 

\myparagraph{Evaluations.} Following previous literature~\cite{dettmers2022llmint8, xiao2022smoothquant, frantar2022gptq, dettmers2022case, zeroquant}, we mainly profiled the quantized models on language modeling tasks (perplexity evaluation on WikiText-2~\cite{merity2016pointer}) since perplexity can stably reflect the LLM's performance~\cite{dettmers2022case}.

\myparagraph{Baselines.} Our primary baseline is vanilla round-to-nearest quantization (RTN). It is actually quite strong when using a small group size like 128~\cite{frantar2022gptq, dettmers2022case}. We also compare with a state-of-the-art method GPTQ~\cite{frantar2022gptq} for LLM weight quantization. For GPTQ, we also compare with an updated version that uses a ``reorder'' trick (denoted as GPTQ-Reorder or GPTQ-R). Other techniques like ZeroQuant~\cite{zeroquant}, AdaRound~\cite{nagel2020up}, and BRECQ~\cite{li2021brecq} rely on backpropagation to update the quantized weights, which may not easily scale up to large model sizes; they also do not outperform GPTQ~\cite{frantar2022gptq}, thus not included for study.

\subsection{Evaluation}
\definecolor{winblue}{RGB}{114, 147, 182}

\begin{figure}%
    \centering
     \includegraphics[width=0.5\textwidth]{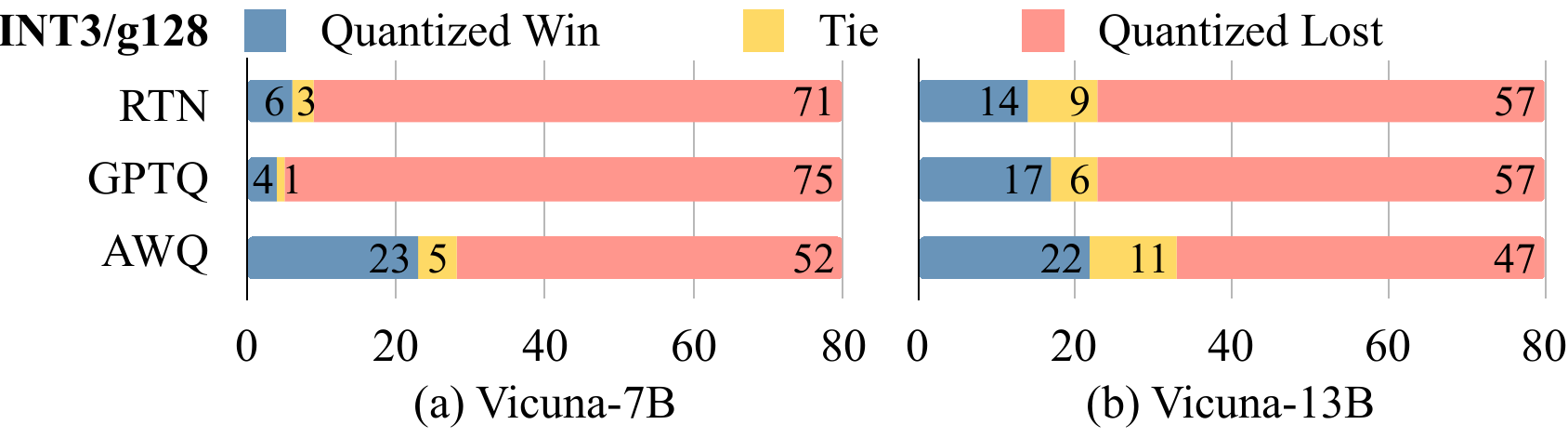}
    \caption{Comparing INT3-g128 quantized Vicuna models with FP16 counterparts under GPT-4 evaluation protocol~\cite{vicuna2023}. More winning cases (in 
    \textcolor{winblue}{blue}) indicate better performance. \methodshort consistently improves the quantized performance compared to RTN and GPTQ~\cite{frantar2022gptq}, showing generalization to instruction-tuned models.
    } 
    \label{fig:vicuna_gpt4_eval}
\end{figure}

\begin{table*}%
    \small
    \setlength{\tabcolsep}{11pt}
    
    \centering
    \begin{tabular}{llcccccc}
        \toprule
        \multicolumn{2}{l}{\textbf{COCO (CIDEr $\uparrow$)}} & 0-shot & 4-shot & 8-shot & 16-shot & 32-shot & \emph{$\Delta$(32-shot)} \\  \midrule
    FP16 & - & 63.73 & 72.18 & 76.95 & 79.74 & 81.70  & - \\ \midrule
    \multirow{3}{*}{\shortstack{INT4\\g128}} & RTN & 60.24 & 68.07 & 72.46 & 74.09 & 77.13 & -4.57 \\ 
    & GPTQ & 59.72 & 67.68 & 72.53 & 74.98 & 74.98 & -6.72 \\ %
    & \methodshort & \textbf{62.57} & \textbf{71.02} & \textbf{74.75} & \textbf{78.23} & \textbf{80.53} & \textbf{-1.17} \\ \midrule
     \multirow{3}{*}{\shortstack{INT3\\g128}} & RTN & 46.07 & 55.13 & 60.46 & 63.21 & 64.79 & -16.91\\ 
    & GPTQ & 29.84 & 50.77 & 56.55 & 60.54 & 64.77 & -16.93\\ %
    & \methodshort & \textbf{56.33} & \textbf{64.73} & \textbf{68.79} & \textbf{72.86} & \textbf{74.47} & \textbf{-7.23}\\
        \bottomrule
    \end{tabular}
    \caption {Quantization results of a visual language model OpenFlamingo-9B~\cite{openflamingo} on COCO Captioning datasets. \method{} outperforms existing methods under zero-shot and various few-shot settings, demonstrating the generability to different modalities and in-context learning workloads. \method{} reduces the quantization degradation (32-shot) from 4.57 to 1.17 under INT4-g128, providing 4$\times$ model size reduction with negligible performance loss. 
    }
    \label{tab:openflamingo}
\end{table*}

\begin{table*}
    \setlength{\tabcolsep}{5pt}
    \small
    \centering
\begin{tabular}{@{}lccccccccccc@{}}
\toprule
\textbf{Model (Accuracy$\uparrow$)} & VQAv2 & GQA & VizWiz & SQA-I & VQA-T & POPE & MME & MMB & SEED & llava-bench & MM-Vet \\
\midrule
VILA-7B           & 80.3 & 63.1 & 59.6   & 68.0  & 62.6  & 86.3  & 1489.4 & 69.8	   & 61.7 & 75.2        & 35.1   \\
VILA-7B-AWQ       & 80.1 & 63.0 & 57.8   & 68.0  & 61.9  & 85.3  & 1486.3 & 68.8   & 61.3 & 75.8        & 35.9   \\
VILA-13B          & 80.5 & 63.6 & 63.1   & 70.5  & 64.0  & 86.3  & 1553.6 & 73.8   & 62.8 & 78.3        & 42.6   \\
VILA-13B-AWQ      & 80.4 & 63.6 & 63.0   & 71.2  & 63.5  & 87.0  & 1552.9 & 73.6   & 62.2 & 77.6        & 42.0   \\
\bottomrule
\end{tabular}
\caption{INT4-g128 results of VILA-7B and VILA-13B~\cite{lin2023vila} on 11 visual-language benchmarks. \methodshort consistently shows lossless performance on all benchmarks. Benchmark names are abbreviated due to space limits. VQA-v2~\cite{goyal2017vqav2}; GQA~\cite{hudson2019gqa}; VisWiz~\cite{gurari2018vizwiz}; SQA$^\text{I}$: ScienceQA-IMG~\cite{lu2022learn}; VQA$^\text{T}$: TextVQA~\cite{singh2019textvqa}; POPE~\cite{li2023pope}; MME~\cite{fu2023mme}; MMB: MMBench~\cite{liu2023mmbench}; MMB$^\text{CN}$: MMBench-Chinese~\cite{liu2023mmbench}; SEED: SEED-Bench~\cite{li2023seed}; LLaVA$^\text{W}$: LLaVA-Bench (In-the-Wild)~\cite{liu2023llava}; MM-Vet~\cite{yu2023mmvet}.}
\label{tab:vila_acc}

\end{table*}

\myparagraph{Results on LLaMA models.}
We focus on LLaMA models (LLaMA~\cite{touvron2023llama} and Llama-2~\cite{touvron2023llama2}) due to their superior performance compared to other open-source LLMs~\cite{opt, scao2022bloom}; it is also the foundation of many popular open-source models~\cite{alpaca, vicuna2023}. We evaluate the perplexity before and after quantization in Table~\ref{tab:llama_opt_ppl}. 
\methodshort consistently outperforms round-to-nearest (RTN) and GPTQ~\cite{frantar2022gptq} (w/ and w/o reordering) across different model scales (7B-70B) and generations. 

\myparagraph{Results on Mistral / Mixtral models.}
We also evaluated \methodshort on the Mistral and Mixtral models, which are among the most popular open-source LLMs and Mixture-of-Experts (MoE) models, respectively~\cite{jiang2023mistral, jiang2024mixtral}. The results indicate that \methodshort achieves superior performance on both the Mistral and Mixtral models. This demonstrates that \methodshort is effective across various model architectures.

\begin{figure*}[t]
    \centering
     \includegraphics[width=\textwidth]{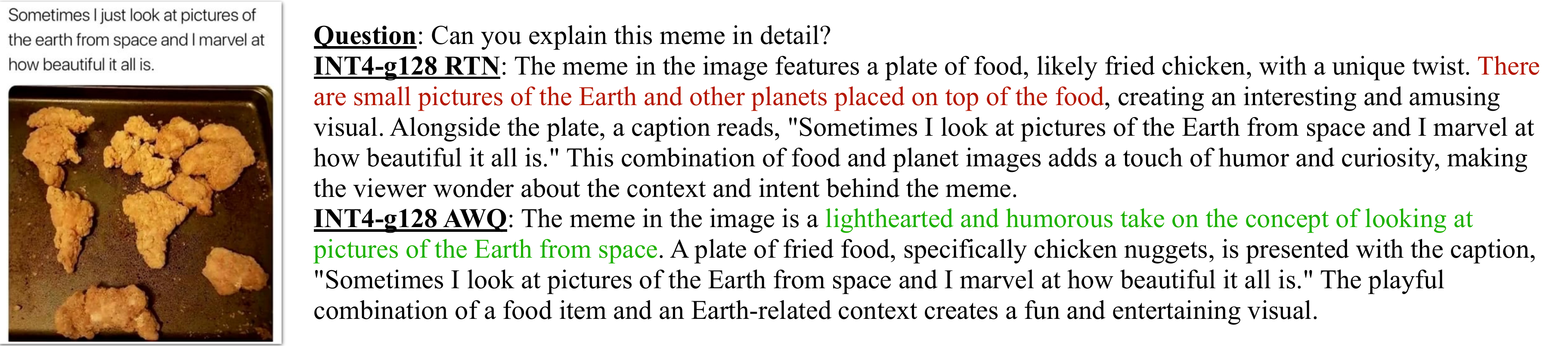}
    \caption{Visual reasoning examples from LLaVA-13B model~\cite{liu2023llava}. \methodshort improves over the round-to-nearest (RTN) baseline, providing more reasonable answers. We color the text to show the \textcolor{codegreen}{correct} or \textcolor{red}{wrong} responses. 
    } 
    \label{fig:visual_reasoning}
\end{figure*}

\begin{figure*}[t]
    \centering
     \includegraphics[width=\textwidth]{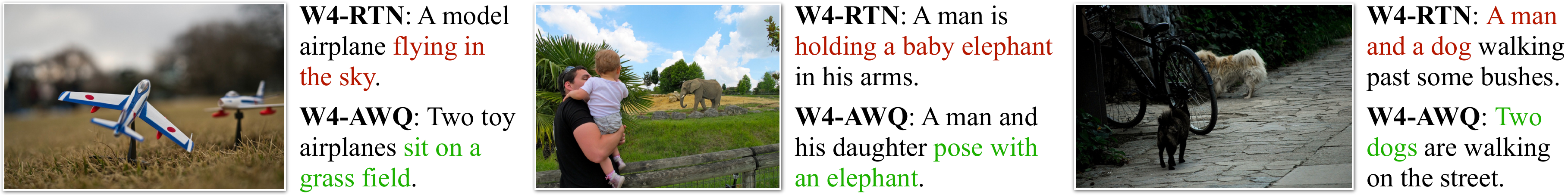}
    \caption{Qualitative results of quantized OpenFlamingo-9B~\cite{openflamingo} on COCO captioning dataset (4-shot, INT4-g128 quantization). Our method significantly improves the captioning quality compared to the round-to-nearest (RTN) baseline. We color the text to show the \textcolor{codegreen}{correct} or \textcolor{red}{wrong} captions. }
    \label{fig:coco_sample}
\end{figure*}

\myparagraph{Quantization of instruction-tuned models.}
Instruction tuning can significantly improve the models' performance and usability 
~\cite{wei2021finetuned,sanh2021multitask,ouyang2022training,chung2022scaling}. It has become an essential procedure before model deployment. We further benchmark our method's performance on a popular instruction-tuned model Vicuna~\cite{vicuna2023} in Figure~\ref{fig:vicuna_gpt4_eval}. We used the GPT-4 score to evaluate the quantized models' performance against the FP16 counterpart on 80 sample questions~\cite{vicuna2023}.
We compare the responses with both orders (quantized-FP16, FP16-quantized) to get rid of the ordering effect (we found GPT-4 tends to increase the rating of the first input), leading to 160 trials. 
\methodshort consistently improves the INT3-g128 quantized Vicuna models over RTN and GPTQ under both scales (7B and 13B), demonstrating the generability to instruction-tuned models.

\begin{table}
    \small
    \setlength{\tabcolsep}{2pt}
    
    \centering
    \begin{tabular}{llcccccc}
        \toprule
        \textbf{MBPP (7B)} & pass@1 & pass@10 \\  \midrule
    FP16 & 38.53 & 49.77 \\ \midrule
    RTN & 37.51 & 48.49  \\ 
    GPTQ & 31.97 & 44.75  \\ %
    \methodshort & \textbf{40.64} & \textbf{49.25}  \\ 
        \bottomrule
    \end{tabular}
    \begin{tabular}{llcccccc}
        \toprule
        \textbf{GSM8K} & 7B & 13B & 70B \\  \midrule
    FP16  & 13.87 & 26.16 & 56.41 \\ \midrule
    RTN & 11.07 & 21.23 & 53.98 \\ 
    GPTQ & 12.13 & 24.26 & 56.03  \\ %
    \methodshort & \textbf{13.57} & \textbf{25.25} & \textbf{56.40} \\ 
        \bottomrule
    \end{tabular}
    \caption {INT4-g128 quantization results of CodeLlama-7b-Instruct-hf on MBPP dataset and Llama-2 (7B/13B/70B) on GSM8K dataset. AWQ outperforms existing methods on programming and math datasets, demonstrating the generability to different scenarios and evaluation settings. Notably, AWQ under the INT4-g128 configuration demonstrates comparable performance to the original FP16 model across both datasets.}
    \label{tab:code_and_math}
\end{table}

\myparagraph{Quantization of multi-modal language models.} Large multi-modal models (LMMs) or visual language models (VLMs) are LLMs augmented with vision inputs~\cite{alayrac2022flamingo,li2023blip,koh2023grounding,driess2023palm,zhang2023llama,liu2023llava}. Such models are able to perform text generation conditioned on image/video inputs. Since our method does not have the overfitting issue to the calibration set, it can be directly applied to VLMs to provide accurate and efficient quantization. 
We perform experiments with the OpenFlamingo-9B model~\cite{openflamingo} (an open-source reproduction of~\cite{alayrac2022flamingo}) on COCO captioning~\cite{chen2015microsoft} dataset (Table~\ref{tab:openflamingo}). We measured the average performance of 5k samples under different few-shot settings. We only quantize the language part of the model since it dominates the model size. 
\methodshort outperforms existing methods under zero-shot and various few-shot settings, demonstrating the generability to different modalities and in-context learning workloads. It reduces the quantization degradation (32-shot) from 4.57 to 1.17 under INT4-g128, providing 4$\times$ model size reduction with negligible performance loss. 
To further demonstrate the generability of \methodshort, we also evaluated \methodshort on one of the SoTA multi-image visual language models: VILA. The result in Table~\ref{tab:vila_acc} shows that \methodshort achieves lossless quantization performance on 11 visual-language benchmarks.
We further provide some qualitative captioning results in Figure~\ref{fig:coco_sample} to show our advantage over RTN.  
Our method provides a push-the-button solution for LMM/VLM quantization. It is the \emph{first} study of VLM low-bit quantization to the best of our knowledge.

\begin{figure*}
    \centering
     \includegraphics[width=\textwidth]{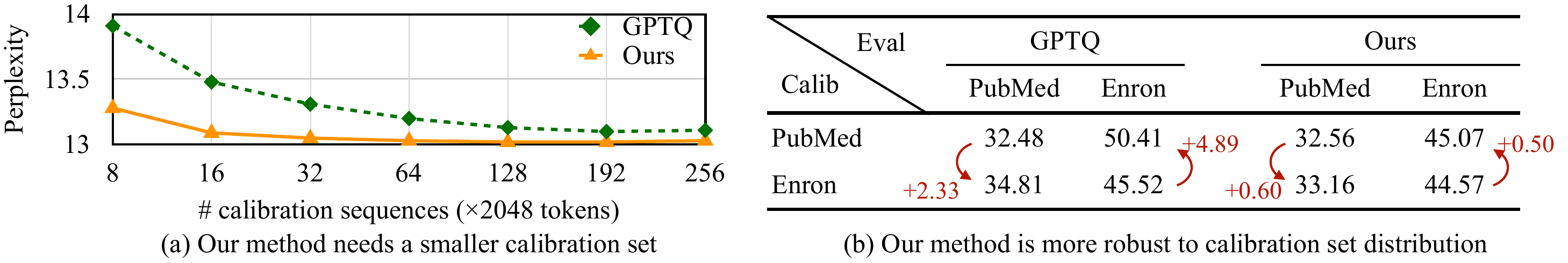}
    \caption{\textbf{Left:} \methodshort needs a much smaller calibration set to reach a good quantized performance. It can achieve better perplexity using 10$\times$ smaller calibration set compared to GPTQ. \textbf{Right:} Our method is more robust to the calibration set distribution. Overall, using the same calibration and evaluation distribution works the best (PubMed-PubMed, Enron-Enron). But when using a different calibration distribution (PubMed-Enron, Enron-PubMed), \methodshort only increases the perplexity by 0.5-0.6, while GPTQ has 2.3-4.9 worse perplexity. All experiments are done with the OPT-6.7B model under INT3-g128 quantization.}
    \label{fig:calib_set_ablation}
\end{figure*}

\begin{figure*}[t]
    \centering
     \includegraphics[width=\textwidth]{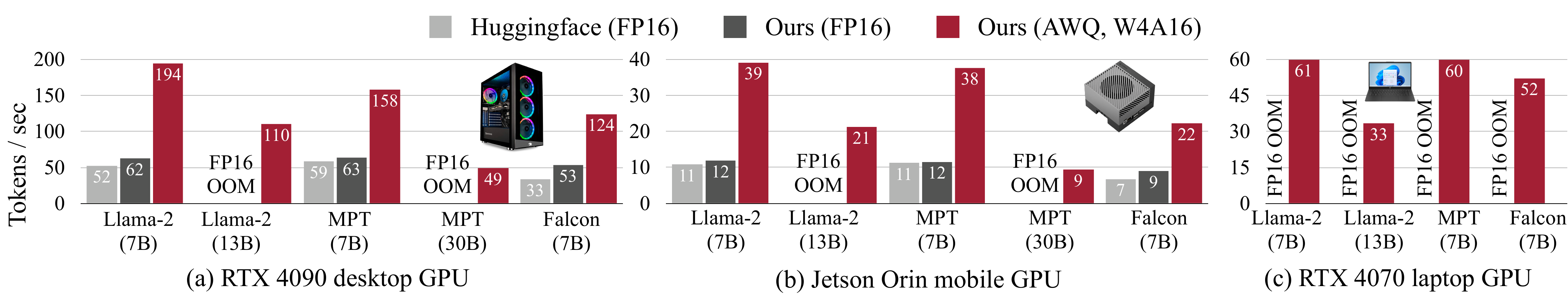}
\caption{\system provides a turn-key solution to transform the theoretical memory footprint reduction into a quantifiable speedup. As a result, \system is up to \textbf{3.9$\times$} and \textbf{3.5$\times$} faster than the FP16 implementation from Huggingface on 4090 (desktop GPU) and Orin (mobile GPU), respectively. AWQ also democratizes Llama-2-13B deployment on laptop GPUs (4070) with merely 8GB memory.} %
    \label{fig:kernel_speedup}
\end{figure*}

\begin{figure*}[t]
    \centering
     \includegraphics[width=\textwidth]{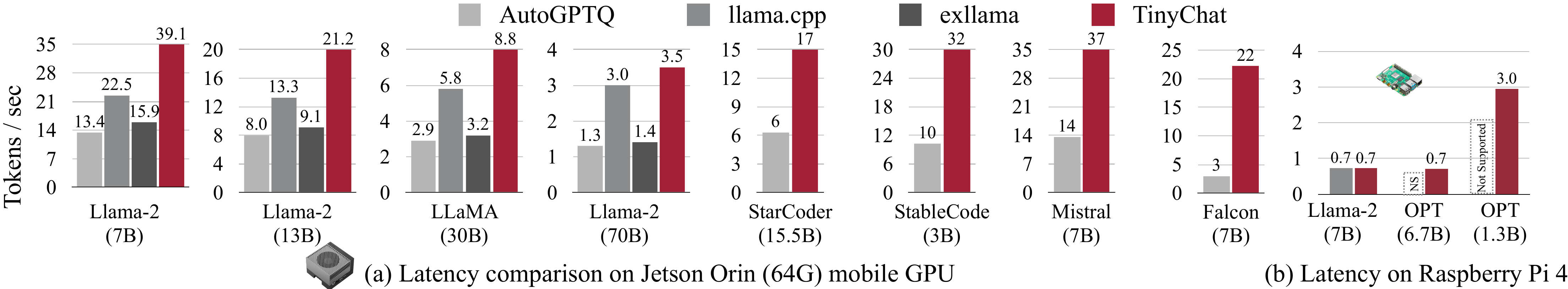}
\caption{\system offers \textbf{1.2-3.0$\times$} speedup over existing systems when running 4-bit quantized Llama models on NVIDIA Jetson Orin. It also supports a diverse range of general-purpose and coding-specific LLMs with at least \textbf{2.6$\times$} speedup over AutoGPTQ, which also supports all these workloads. Moreover, \system seamlessly operates on Raspberry Pi and enables the deployment of LLMs with up to 7 billion parameters on extremely resource-constrained IoT devices.} %
    \label{fig:kernel_speedup_comparisons}
\end{figure*}

\begin{table}
\small
    \setlength{\tabcolsep}{3pt}
    
    \centering
    \begin{tabular}{llccccc}
        \toprule
          \textbf{OPT (Wiki PPL$\downarrow$)} & 1.3B & 2.7B & 6.7B & 13B & 30B  \\  \midrule
        FP16& 14.62 & 12.47 & 10.86 & 10.13 & 9.56 \\ \midrule
        RTN & 10476 & 193210 & 7622 & 17564 & 8170 \\
        GPTQ & 46.67 & 28.15 & 16.65 & 16.74 &  11.75 \\ \cmidrule(lr){2-6}
        \methodshort+GPTQ & \textbf{35.71} & \textbf{25.70} & \textbf{15.71} & \textbf{13.25} & \textbf{11.38}   \\
        \bottomrule
    \end{tabular}
    \caption{Our method is orthogonal to GPTQ: it further closes the performance gap under extreme low-bit quantization (INT2-g64) when combined with GPTQ. Results are WikiText-2 perplexity of OPT models.  }
    \label{tab:int2}
\end{table}

\myparagraph{Visual reasoning results.}
We further provide some qualitative visual reasoning examples of the LLaVA-13B~\cite{liu2023llava} model in Figure~\ref{fig:visual_reasoning}. \methodshort improves the responses compared to round-to-nearest (RTN) for INT4-g128 quantization, leading to more reasonable answers. In this first example, the \methodshort model can understand the meme as it resembles the Earth when looking from space, while RTN produces wrong descriptions (marked in \textcolor{red}{red}). %

\myparagraph{Results on programming and math tasks}
To further evaluate the performance of AWQ on tasks involving complex generations, we also tested AWQ on MBPP~\cite{austin2021program} and GSM8K~\cite{cobbe2021training}. MBPP~\cite{austin2021program} consists of around 1,000 Python programming problems, designed to be solvable by entry level programmers, covering programming fundamentals, standard library functionality, etc. GSM8K~\cite{cobbe2021training} was created to support the task of question answering on basic mathematical problems that require multi-step reasoning. We quantize CodeLlama-7b-Instruct-hf and Llama-2 to INT4-g128 and perform experiments on programming and math datasets (Table~\ref{tab:code_and_math}). AWQ outperforms existing methods on both datasets, demonstrating the generability to complex generation. AWQ under the INT4-g128 configuration demonstrates comparable
performance to the original FP16 model on both datasets.

\myparagraph{Extreme low-bit quantization.}
We further quantize LLM to INT2 to accommodate limited device memory (Table~\ref{tab:int2}). RTN completely fails, and \methodshort brings significant perplexity improvement on top of GPTQ.%
Our method is orthogonal to GPTQ.
We can combine our method with GPTQ to further improve the INT2 quantization performance, making it a more practical setting. 

\subsection{Data Efficiency and Generalization}
\label{sec:ablation_study}
\myparagraph{Better data-efficiency for the calibration set.} Our method requires a smaller calibration set since we do not rely on regression/backpropagation; we only measure the average activation scale from the calibration set, which is data-efficient.
To demonstrate the idea, we compare the perplexity of the OPT-6.7B model with INT3-g128 quantization in Figure~\ref{fig:calib_set_ablation} (a). \methodshort needs a much smaller calibration to reach a good quantized performance; it can achieve better perplexity using 10$\times$ smaller calibration set compared to GPTQ (16 sequences \emph{v.s.} 192 sequences).

\myparagraph{Robust to the calibration set distributions. }
Our method is less sensitive to the calibration set distribution since we only measure the average activation scale from the calibration set, which is more generalizable across different dataset distributions. We further benchmarked the effect of the different calibration set distributions in Figure~\ref{fig:calib_set_ablation}(b). We took two subsets from the Pile dataset~\cite{gao2020pile}: PubMed Abstracts and Enron Emails~\cite{klimt2004enron}. We use each of the subsets as the calibration set and evaluate the quantized model on both sets (the calibration and evaluation sets are split with no overlapping; we used 1k samples for evaluation). Overall, using the same calibration and evaluation distribution works the best (PubMed-PubMed, Enron-Enron). But when using a different calibration distribution (PubMed-Enron, Enron-PubMed), \methodshort only increases the perplexity by 0.5-0.6, while GPTQ has 2.3-4.9 worse perplexity. This demonstrates the robustness of \methodshort to the calibration set distribution. 

\subsection{Speedup Evaluation}
\begin{table}
    \setlength{\tabcolsep}{5pt}
    \small
    \centering
    \begin{tabular}{lcccc} \toprule
\textbf{Model (Throughput$\uparrow$)}          & Precision & A100 & 4090 & Orin \\ \midrule
VILA-7B      & FP16      & 81.6              & 58.5              & 11.5              \\
VILA-7B-AWQ  & W4A16     & 155.3             & 168.1             & 35.6              \\ \midrule
VILA-13B     & FP16      & 48.5              & OOM               & 6.1               \\
VILA-13B-AWQ & W4A16     & 102.1             & 99.0              & 17.5 \\\bottomrule             
\end{tabular}
    \caption{\system also enables seamless deployment of VILA~\cite{lin2023vila}, a state-of-the-art visual-language model, on multiple GPU platforms. Leveraging our 4-bit AWQ quantization, \system accelerates VILA-7B by up to \textbf{3.1$\times$} and VILA-13B by up to \textbf{2.9$\times$}.}
    \label{tab:vila_latency}
\end{table}

\myparagraph{Settings.} In Figure~\ref{fig:kernel_speedup}, we demonstrate the system acceleration results from \system. \system optimizes both linear layers and layers that do not have quantized weights. We conduct benchmarking experiments on RTX 4090 and Jetson Orin following the protocol described in exllama~\footnote{\url{https://github.com/turboderp/exllama}}. We perform batch size = 1 inference for all LLMs using a fixed prompt length of 4 tokens. We generate 200 tokens for each inference run and calculate the median latency as the final result.

\myparagraph{Results.} As in Figure~\ref{fig:kernel_speedup}(a), \system brings \textbf{2.7-3.9$\times$} speedup to three families of LLMs (Llama-2, MPT and Falcon) on 4090 compared with the Huggingface FP16 implementation. For Llama-2-7B, we improve the inference speed from 52 tokens/s to 62 tokens/s through FP16 kernel fusion. On top of the stronger FP16 baseline, we further harvest \textbf{3.1$\times$} additional speedup from the fast quantized linear kernels. For Falcon-7B, the official implementation did not support KV cache correctly during the inference time, and thus it is significantly slower than other models. In this case, our FP16 optimizations bring about a larger speedup of \textbf{1.6$\times$}. On the laptop 4070 GPU with only 8GB memory, we are still able to run Llama-2-13B models at 33 tokens/s, while the FP16 implementation cannot fit 7B models. We also demonstrate visual-language model~\cite{lin2023vila} acceleration results in Table \ref{tab:vila_latency}. \system brings about \textbf{3$\times$} speedup to both VILA-7B and VILA-13B on NVIDIA Jetson Orin. Notably, we implement the forward pass for all AWQ models using native PyTorch APIs, and this code is reused across various GPU architectures. Hence, \system offers exceptional extensibility. %

\myparagraph{Comparisons against other systems.}
We compare \system against existing edge LLM inference systems AutoGPTQ, llama.cpp and exllama in Figure~\ref{fig:kernel_speedup_comparisons}. Our system achieves up to 1.7$\times$ speedup over llama.cpp on Orin. Furthermore, llama.cpp and exllama exhibit limited adaptability, primarily tailored for LLaMA and Llama-2 models. In contrast, our \system supports a wide range of applications, including StarCoder~\cite{li2023starcoder}, StableCode (GPT-NeoX)~\cite{black2022gpt}, Mistral~\cite{jiang2023mistral}, and Falcon~\cite{penedo2023refinedweb} while consistently delivering significant speedup over AutoGPTQ. \system even democratizes LLM deployment on extremely resource-constrained Raspberry Pi 4B, achieving 0.7 tokens/s for 7B models.

\section{Conclusion}
In this work, we propose \method (\methodshort), a simple yet effective method for low-bit weight-only LLM compression.
Based on the observation that weights are not equally important in LLMs, \methodshort performs per-channel scaling to reduce the quantization loss of salient weights. \methodshort does not over-fit the calibration set and preserves the generalist abilities of LLMs in various domains and modalities. It outperforms existing work on language modeling and is applicable to instruction-tuned LMs and multi-modal LMs. Our \system system further translates the
theoretical memory savings achieved by \methodshort into \textbf{3.2-3.3$\times$} measured speedups over the FP16 implementations from Huggingface on desktop and mobile GPUs, democratizing LLM deployment on the edge.

\section*{Acknowledgements}
We thank MIT AI Hardware Program, National Science Foundation (CNS-2112562), MIT-IBM Watson AI Lab, Amazon and MIT Science Hub, Microsoft Turing Academic Program, and Samsung for supporting this research.

\bibliography{main}

\begin{thebibliography}{71}
\providecommand{\natexlab}[1]{#1}
\providecommand{\url}[1]{\texttt{#1}}
\expandafter\ifx\csname urlstyle\endcsname\relax
  \providecommand{\doi}[1]{doi: #1}\else
  \providecommand{\doi}{doi: \begingroup \urlstyle{rm}\Url}\fi

\bibitem[Alayrac et~al.(2022)Alayrac, Donahue, Luc, Miech, Barr, Hasson, Lenc, Mensch, Millican, Reynolds, et~al.]{alayrac2022flamingo}
Alayrac, J.-B., Donahue, J., Luc, P., Miech, A., Barr, I., Hasson, Y., Lenc, K., Mensch, A., Millican, K., Reynolds, M., et~al.
\newblock Flamingo: a visual language model for few-shot learning.
\newblock \emph{Advances in Neural Information Processing Systems}, 35:\penalty0 23716--23736, 2022.

\bibitem[Austin et~al.(2021)Austin, Odena, Nye, Bosma, Michalewski, Dohan, Jiang, Cai, Terry, Le, and Sutton]{austin2021program}
Austin, J., Odena, A., Nye, M., Bosma, M., Michalewski, H., Dohan, D., Jiang, E., Cai, C., Terry, M., Le, Q., and Sutton, C.
\newblock Program synthesis with large language models, 2021.

\bibitem[Awadalla et~al.(2023)Awadalla, Gao, Gardner, Hessel, Hanafy, Zhu, Marathe, Bitton, Gadre, Jitsev, Kornblith, Koh, Ilharco, Wortsman, and Schmidt]{openflamingo}
Awadalla, A., Gao, I., Gardner, J., Hessel, J., Hanafy, Y., Zhu, W., Marathe, K., Bitton, Y., Gadre, S., Jitsev, J., Kornblith, S., Koh, P.~W., Ilharco, G., Wortsman, M., and Schmidt, L.
\newblock Openflamingo, March 2023.
\newblock URL \url{https://doi.org/10.5281/zenodo.7733589}.

\bibitem[Bengio et~al.(2013)Bengio, L{\'e}onard, and Courville]{bengio2013estimating}
Bengio, Y., L{\'e}onard, N., and Courville, A.
\newblock Estimating or propagating gradients through stochastic neurons for conditional computation.
\newblock \emph{arXiv preprint arXiv:1308.3432}, 2013.

\bibitem[Black et~al.(2022)Black, Biderman, Hallahan, Anthony, Gao, Golding, He, Leahy, McDonell, Phang, et~al.]{black2022gpt}
Black, S., Biderman, S., Hallahan, E., Anthony, Q., Gao, L., Golding, L., He, H., Leahy, C., McDonell, K., Phang, J., et~al.
\newblock Gpt-neox-20b: An open-source autoregressive language model.
\newblock \emph{arXiv preprint arXiv:2204.06745}, 2022.

\bibitem[Brown et~al.(2020)Brown, Mann, Ryder, Subbiah, Kaplan, Dhariwal, Neelakantan, Shyam, Sastry, Askell, Agarwal, Herbert-Voss, Krueger, Henighan, Child, Ramesh, Ziegler, Wu, Winter, Hesse, Chen, Sigler, Litwin, Gray, Chess, Clark, Berner, McCandlish, Radford, Sutskever, and Amodei]{gpt3}
Brown, T., Mann, B., Ryder, N., Subbiah, M., Kaplan, J.~D., Dhariwal, P., Neelakantan, A., Shyam, P., Sastry, G., Askell, A., Agarwal, S., Herbert-Voss, A., Krueger, G., Henighan, T., Child, R., Ramesh, A., Ziegler, D., Wu, J., Winter, C., Hesse, C., Chen, M., Sigler, E., Litwin, M., Gray, S., Chess, B., Clark, J., Berner, C., McCandlish, S., Radford, A., Sutskever, I., and Amodei, D.
\newblock Language models are few-shot learners.
\newblock In Larochelle, H., Ranzato, M., Hadsell, R., Balcan, M., and Lin, H. (eds.), \emph{Advances in Neural Information Processing Systems}, volume~33, pp.\  1877--1901. Curran Associates, Inc., 2020.
\newblock URL \url{https://proceedings.neurips.cc/paper/2020/file/1457c0d6bfcb4967418bfb8ac142f64a-Paper.pdf}.

\bibitem[Chen et~al.(2018)Chen, Moreau, Jiang, Zheng, Yan, Shen, Cowan, Wang, Hu, Ceze, et~al.]{chen2018tvm}
Chen, T., Moreau, T., Jiang, Z., Zheng, L., Yan, E., Shen, H., Cowan, M., Wang, L., Hu, Y., Ceze, L., et~al.
\newblock {TVM: An Automated End-to-End Optimizing Compiler for Deep Learning}.
\newblock In \emph{13th USENIX Symposium on Operating Systems Design and Implementation (OSDI)}, 2018.

\bibitem[Chen et~al.(2015)Chen, Fang, Lin, Vedantam, Gupta, Doll{\'a}r, and Zitnick]{chen2015microsoft}
Chen, X., Fang, H., Lin, T.-Y., Vedantam, R., Gupta, S., Doll{\'a}r, P., and Zitnick, C.~L.
\newblock Microsoft coco captions: Data collection and evaluation server.
\newblock \emph{arXiv preprint arXiv:1504.00325}, 2015.

\bibitem[Chiang et~al.(2023)Chiang, Li, Lin, Sheng, Wu, Zhang, Zheng, Zhuang, Zhuang, Gonzalez, Stoica, and Xing]{vicuna2023}
Chiang, W.-L., Li, Z., Lin, Z., Sheng, Y., Wu, Z., Zhang, H., Zheng, L., Zhuang, S., Zhuang, Y., Gonzalez, J.~E., Stoica, I., and Xing, E.~P.
\newblock Vicuna: An open-source chatbot impressing gpt-4 with 90\%* chatgpt quality, March 2023.
\newblock URL \url{https://lmsys.org/blog/2023-03-30-vicuna/}.

\bibitem[Choi et~al.(2018)Choi, Wang, Venkataramani, Chuang, Srinivasan, and Gopalakrishnan]{choi2018pact}
Choi, J., Wang, Z., Venkataramani, S., Chuang, P. I.-J., Srinivasan, V., and Gopalakrishnan, K.
\newblock Pact: Parameterized clipping activation for quantized neural networks.
\newblock \emph{arXiv preprint arXiv:1805.06085}, 2018.

\bibitem[Chung et~al.(2022)Chung, Hou, Longpre, Zoph, Tay, Fedus, Li, Wang, Dehghani, Brahma, et~al.]{chung2022scaling}
Chung, H.~W., Hou, L., Longpre, S., Zoph, B., Tay, Y., Fedus, W., Li, E., Wang, X., Dehghani, M., Brahma, S., et~al.
\newblock Scaling instruction-finetuned language models.
\newblock \emph{arXiv preprint arXiv:2210.11416}, 2022.

\bibitem[Cobbe et~al.(2021)Cobbe, Kosaraju, Bavarian, Chen, Jun, Kaiser, Plappert, Tworek, Hilton, Nakano, Hesse, and Schulman]{cobbe2021training}
Cobbe, K., Kosaraju, V., Bavarian, M., Chen, M., Jun, H., Kaiser, L., Plappert, M., Tworek, J., Hilton, J., Nakano, R., Hesse, C., and Schulman, J.
\newblock Training verifiers to solve math word problems, 2021.

\bibitem[Dettmers \& Zettlemoyer(2022)Dettmers and Zettlemoyer]{dettmers2022case}
Dettmers, T. and Zettlemoyer, L.
\newblock The case for 4-bit precision: k-bit inference scaling laws.
\newblock \emph{arXiv preprint arXiv:2212.09720}, 2022.

\bibitem[Dettmers et~al.(2022)Dettmers, Lewis, Belkada, and Zettlemoyer]{dettmers2022llmint8}
Dettmers, T., Lewis, M., Belkada, Y., and Zettlemoyer, L.
\newblock Llm.int8(): 8-bit matrix multiplication for transformers at scale.
\newblock \emph{arXiv preprint arXiv:2208.07339}, 2022.

\bibitem[Driess et~al.(2023)Driess, Xia, Sajjadi, Lynch, Chowdhery, Ichter, Wahid, Tompson, Vuong, Yu, et~al.]{driess2023palm}
Driess, D., Xia, F., Sajjadi, M.~S., Lynch, C., Chowdhery, A., Ichter, B., Wahid, A., Tompson, J., Vuong, Q., Yu, T., et~al.
\newblock Palm-e: An embodied multimodal language model.
\newblock \emph{arXiv preprint arXiv:2303.03378}, 2023.

\bibitem[Esser et~al.(2019)Esser, McKinstry, Bablani, Appuswamy, and Modha]{esser2019learned}
Esser, S.~K., McKinstry, J.~L., Bablani, D., Appuswamy, R., and Modha, D.~S.
\newblock Learned step size quantization.
\newblock \emph{arXiv preprint arXiv:1902.08153}, 2019.

\bibitem[Feng et~al.(2023)Feng, Hou, Jin, Lin, Shao, Lai, Ye, Zheng, Yu, Yu, and Chen]{feng2022tensorir}
Feng, S., Hou, B., Jin, H., Lin, W., Shao, J., Lai, R., Ye, Z., Zheng, L., Yu, C.~H., Yu, Y., and Chen, T.
\newblock {TensorIR: An Abstraction for Automatic Tensorized Program Optimization}.
\newblock In \emph{ASPLOS}, 2023.

\bibitem[Frankle \& Carbin(2018)Frankle and Carbin]{frankle2018lottery}
Frankle, J. and Carbin, M.
\newblock The lottery ticket hypothesis: Finding sparse, trainable neural networks.
\newblock \emph{arXiv preprint arXiv:1803.03635}, 2018.

\bibitem[Frantar et~al.(2022)Frantar, Ashkboos, Hoefler, and Alistarh]{frantar2022gptq}
Frantar, E., Ashkboos, S., Hoefler, T., and Alistarh, D.
\newblock Gptq: Accurate post-training quantization for generative pre-trained transformers.
\newblock \emph{arXiv preprint arXiv:2210.17323}, 2022.

\bibitem[Fu et~al.(2023)Fu, Chen, Shen, Qin, Zhang, Lin, Yang, Zheng, Li, Sun, Wu, and Ji]{fu2023mme}
Fu, C., Chen, P., Shen, Y., Qin, Y., Zhang, M., Lin, X., Yang, J., Zheng, X., Li, K., Sun, X., Wu, Y., and Ji, R.
\newblock {MME: A Comprehensive Evaluation Benchmark for Multimodal Large Language Models}.
\newblock \emph{arXiv preprint arXiv:2306.13394}, 2023.

\bibitem[Gao et~al.(2020)Gao, Biderman, Black, Golding, Hoppe, Foster, Phang, He, Thite, Nabeshima, et~al.]{gao2020pile}
Gao, L., Biderman, S., Black, S., Golding, L., Hoppe, T., Foster, C., Phang, J., He, H., Thite, A., Nabeshima, N., et~al.
\newblock The pile: An 800gb dataset of diverse text for language modeling.
\newblock \emph{arXiv preprint arXiv:2101.00027}, 2020.

\bibitem[Gholami et~al.(2021)Gholami, Kim, Dong, Yao, Mahoney, and Keutzer]{gholami2021survey}
Gholami, A., Kim, S., Dong, Z., Yao, Z., Mahoney, M.~W., and Keutzer, K.
\newblock A survey of quantization methods for efficient neural network inference.
\newblock \emph{arXiv preprint arXiv:2103.13630}, 2021.

\bibitem[Goyal et~al.(2017)Goyal, Khot, Summers-Stay, Batra, and Parikh]{goyal2017vqav2}
Goyal, Y., Khot, T., Summers-Stay, D., Batra, D., and Parikh, D.
\newblock Making the v in vqa matter: Elevating the role of image understanding in visual question answering.
\newblock In \emph{Proceedings of the IEEE conference on computer vision and pattern recognition}, pp.\  6904--6913, 2017.

\bibitem[Gurari et~al.(2018)Gurari, Li, Stangl, Guo, Lin, Grauman, Luo, and Bigham]{gurari2018vizwiz}
Gurari, D., Li, Q., Stangl, A.~J., Guo, A., Lin, C., Grauman, K., Luo, J., and Bigham, J.~P.
\newblock Vizwiz grand challenge: Answering visual questions from blind people.
\newblock In \emph{Proceedings of the IEEE conference on computer vision and pattern recognition}, pp.\  3608--3617, 2018.

\bibitem[Han et~al.(2015)Han, Pool, Tran, and Dally]{han2015learning}
Han, S., Pool, J., Tran, J., and Dally, W.
\newblock Learning both weights and connections for efficient neural network.
\newblock \emph{Advances in neural information processing systems}, 28, 2015.

\bibitem[Han et~al.(2016)Han, Mao, and Dally]{han2016deep}
Han, S., Mao, H., and Dally, W.~J.
\newblock {Deep Compression: Compressing Deep Neural Networks with Pruning, Trained Quantization and Huffman Coding}.
\newblock In \emph{ICLR}, 2016.

\bibitem[Hudson \& Manning(2019)Hudson and Manning]{hudson2019gqa}
Hudson, D.~A. and Manning, C.~D.
\newblock Gqa: A new dataset for real-world visual reasoning and compositional question answering.
\newblock In \emph{CVPR}, 2019.

\bibitem[Jacob et~al.(2018)Jacob, Kligys, Chen, Zhu, Tang, Howard, Adam, and Kalenichenko]{jacob2018quantization}
Jacob, B., Kligys, S., Chen, B., Zhu, M., Tang, M., Howard, A., Adam, H., and Kalenichenko, D.
\newblock Quantization and training of neural networks for efficient integer-arithmetic-only inference.
\newblock In \emph{Proceedings of the IEEE Conference on Computer Vision and Pattern Recognition}, pp.\  2704--2713, 2018.

\bibitem[Jiang et~al.(2023)Jiang, Sablayrolles, Mensch, Bamford, Chaplot, Casas, Bressand, Lengyel, Lample, Saulnier, et~al.]{jiang2023mistral}
Jiang, A.~Q., Sablayrolles, A., Mensch, A., Bamford, C., Chaplot, D.~S., Casas, D. d.~l., Bressand, F., Lengyel, G., Lample, G., Saulnier, L., et~al.
\newblock Mistral 7b.
\newblock \emph{arXiv preprint arXiv:2310.06825}, 2023.

\bibitem[Jiang et~al.(2024)Jiang, Sablayrolles, Roux, Mensch, Savary, Bamford, Chaplot, de~las Casas, Hanna, Bressand, Lengyel, Bour, Lample, Lavaud, Saulnier, Lachaux, Stock, Subramanian, Yang, Antoniak, Scao, Gervet, Lavril, Wang, Lacroix, and Sayed]{jiang2024mixtral}
Jiang, A.~Q., Sablayrolles, A., Roux, A., Mensch, A., Savary, B., Bamford, C., Chaplot, D.~S., de~las Casas, D., Hanna, E.~B., Bressand, F., Lengyel, G., Bour, G., Lample, G., Lavaud, L.~R., Saulnier, L., Lachaux, M.-A., Stock, P., Subramanian, S., Yang, S., Antoniak, S., Scao, T.~L., Gervet, T., Lavril, T., Wang, T., Lacroix, T., and Sayed, W.~E.
\newblock Mixtral of experts, 2024.

\bibitem[Kim et~al.(2022)Kim, Henry, Fahim, and Awadalla]{kim2022says}
Kim, Y.~J., Henry, R., Fahim, R., and Awadalla, H.~H.
\newblock Who says elephants can't run: Bringing large scale moe models into cloud scale production.
\newblock \emph{arXiv preprint arXiv:2211.10017}, 2022.

\bibitem[Klimt \& Yang(2004)Klimt and Yang]{klimt2004enron}
Klimt, B. and Yang, Y.
\newblock The enron corpus: A new dataset for email classification research.
\newblock In \emph{Machine Learning: ECML 2004: 15th European Conference on Machine Learning, Pisa, Italy, September 20-24, 2004. Proceedings 15}, pp.\  217--226. Springer, 2004.

\bibitem[Koh et~al.(2023)Koh, Salakhutdinov, and Fried]{koh2023grounding}
Koh, J.~Y., Salakhutdinov, R., and Fried, D.
\newblock Grounding language models to images for multimodal generation.
\newblock \emph{arXiv preprint arXiv:2301.13823}, 2023.

\bibitem[Li et~al.(2023{\natexlab{a}})Li, Wang, Wang, Ge, Ge, and Shan]{li2023seed}
Li, B., Wang, R., Wang, G., Ge, Y., Ge, Y., and Shan, Y.
\newblock Seed-bench: Benchmarking multimodal llms with generative comprehension.
\newblock \emph{arXiv preprint arXiv:2307.16125}, 2023{\natexlab{a}}.

\bibitem[Li et~al.(2023{\natexlab{b}})Li, Li, Savarese, and Hoi]{li2023blip}
Li, J., Li, D., Savarese, S., and Hoi, S.
\newblock Blip-2: Bootstrapping language-image pre-training with frozen image encoders and large language models.
\newblock \emph{arXiv preprint arXiv:2301.12597}, 2023{\natexlab{b}}.

\bibitem[Li et~al.(2023{\natexlab{c}})Li, Allal, Zi, Muennighoff, Kocetkov, Mou, Marone, Akiki, Li, Chim, et~al.]{li2023starcoder}
Li, R., Allal, L.~B., Zi, Y., Muennighoff, N., Kocetkov, D., Mou, C., Marone, M., Akiki, C., Li, J., Chim, J., et~al.
\newblock Starcoder: may the source be with you!
\newblock \emph{arXiv preprint arXiv:2305.06161}, 2023{\natexlab{c}}.

\bibitem[Li et~al.(2021)Li, Gong, Tan, Yang, Hu, Zhang, Yu, Wang, and Gu]{li2021brecq}
Li, Y., Gong, R., Tan, X., Yang, Y., Hu, P., Zhang, Q., Yu, F., Wang, W., and Gu, S.
\newblock Brecq: Pushing the limit of post-training quantization by block reconstruction.
\newblock \emph{arXiv preprint arXiv:2102.05426}, 2021.

\bibitem[Li et~al.(2023{\natexlab{d}})Li, Du, Zhou, Wang, Zhao, and Wen]{li2023pope}
Li, Y., Du, Y., Zhou, K., Wang, J., Zhao, W.~X., and Wen, J.-R.
\newblock Evaluating object hallucination in large vision-language models.
\newblock \emph{arXiv preprint arXiv:2305.10355}, 2023{\natexlab{d}}.

\bibitem[Lin et~al.(2020)Lin, Chen, Lin, Gan, Han, et~al.]{lin2020mcunet}
Lin, J., Chen, W.-M., Lin, Y., Gan, C., Han, S., et~al.
\newblock Mcunet: Tiny deep learning on iot devices.
\newblock \emph{Advances in Neural Information Processing Systems}, 33:\penalty0 11711--11722, 2020.

\bibitem[Lin et~al.(2024)Lin, Yin, Ping, Lu, Molchanov, Tao, Mao, Kautz, Shoeybi, and Han]{lin2023vila}
Lin, J., Yin, H., Ping, W., Lu, Y., Molchanov, P., Tao, A., Mao, H., Kautz, J., Shoeybi, M., and Han, S.
\newblock Vila: On pre-training for visual language models.
\newblock In \emph{CVPR}, 2024.

\bibitem[Liu et~al.(2023{\natexlab{a}})Liu, Li, Wu, and Lee]{liu2023llava}
Liu, H., Li, C., Wu, Q., and Lee, Y.~J.
\newblock Visual instruction tuning.
\newblock 2023{\natexlab{a}}.

\bibitem[Liu et~al.(2023{\natexlab{b}})Liu, Duan, Zhang, Li, Zhang, Zhao, Yuan, Wang, He, Liu, et~al.]{liu2023mmbench}
Liu, Y., Duan, H., Zhang, Y., Li, B., Zhang, S., Zhao, W., Yuan, Y., Wang, J., He, C., Liu, Z., et~al.
\newblock Mmbench: Is your multi-modal model an all-around player?
\newblock \emph{arXiv preprint arXiv:2307.06281}, 2023{\natexlab{b}}.

\bibitem[Lu et~al.(2022)Lu, Mishra, Xia, Qiu, Chang, Zhu, Tafjord, Clark, and Kalyan]{lu2022learn}
Lu, P., Mishra, S., Xia, T., Qiu, L., Chang, K.-W., Zhu, S.-C., Tafjord, O., Clark, P., and Kalyan, A.
\newblock Learn to explain: Multimodal reasoning via thought chains for science question answering.
\newblock \emph{Advances in Neural Information Processing Systems}, 35:\penalty0 2507--2521, 2022.

\bibitem[Merity et~al.(2016)Merity, Xiong, Bradbury, and Socher]{merity2016pointer}
Merity, S., Xiong, C., Bradbury, J., and Socher, R.
\newblock Pointer sentinel mixture models, 2016.

\bibitem[MLC-Team(2023)]{mlc-llm}
MLC-Team.
\newblock {MLC-LLM}, 2023.
\newblock URL \url{https://github.com/mlc-ai/mlc-llm}.

\bibitem[Nagel et~al.(2019)Nagel, Baalen, Blankevoort, and Welling]{nagel2019data}
Nagel, M., Baalen, M.~v., Blankevoort, T., and Welling, M.
\newblock Data-free quantization through weight equalization and bias correction.
\newblock In \emph{Proceedings of the IEEE/CVF International Conference on Computer Vision}, pp.\  1325--1334, 2019.

\bibitem[Nagel et~al.(2020)Nagel, Amjad, Van~Baalen, Louizos, and Blankevoort]{nagel2020up}
Nagel, M., Amjad, R.~A., Van~Baalen, M., Louizos, C., and Blankevoort, T.
\newblock Up or down? adaptive rounding for post-training quantization.
\newblock In \emph{International Conference on Machine Learning}, pp.\  7197--7206. PMLR, 2020.

\bibitem[Nagel et~al.(2021)Nagel, Fournarakis, Amjad, Bondarenko, Van~Baalen, and Blankevoort]{nagel2021white}
Nagel, M., Fournarakis, M., Amjad, R.~A., Bondarenko, Y., Van~Baalen, M., and Blankevoort, T.
\newblock A white paper on neural network quantization.
\newblock \emph{arXiv preprint arXiv:2106.08295}, 2021.

\bibitem[Ouyang et~al.(2022)Ouyang, Wu, Jiang, Almeida, Wainwright, Mishkin, Zhang, Agarwal, Slama, Ray, et~al.]{ouyang2022training}
Ouyang, L., Wu, J., Jiang, X., Almeida, D., Wainwright, C., Mishkin, P., Zhang, C., Agarwal, S., Slama, K., Ray, A., et~al.
\newblock Training language models to follow instructions with human feedback.
\newblock \emph{Advances in Neural Information Processing Systems}, 35:\penalty0 27730--27744, 2022.

\bibitem[Park et~al.(2022)Park, Park, Kwon, Kim, Lee, and Lee]{park2022nuqmm}
Park, G., Park, B., Kwon, S.~J., Kim, B., Lee, Y., and Lee, D.
\newblock nuqmm: Quantized matmul for efficient inference of large-scale generative language models.
\newblock \emph{arXiv preprint arXiv:2206.09557}, 2022.

\bibitem[Penedo et~al.(2023)Penedo, Malartic, Hesslow, Cojocaru, Cappelli, Alobeidli, Pannier, Almazrouei, and Launay]{penedo2023refinedweb}
Penedo, G., Malartic, Q., Hesslow, D., Cojocaru, R., Cappelli, A., Alobeidli, H., Pannier, B., Almazrouei, E., and Launay, J.
\newblock The refinedweb dataset for falcon llm: outperforming curated corpora with web data, and web data only.
\newblock \emph{arXiv preprint arXiv:2306.01116}, 2023.

\bibitem[Sanh et~al.(2021)Sanh, Webson, Raffel, Bach, Sutawika, Alyafeai, Chaffin, Stiegler, Scao, Raja, et~al.]{sanh2021multitask}
Sanh, V., Webson, A., Raffel, C., Bach, S.~H., Sutawika, L., Alyafeai, Z., Chaffin, A., Stiegler, A., Scao, T.~L., Raja, A., et~al.
\newblock Multitask prompted training enables zero-shot task generalization.
\newblock \emph{arXiv preprint arXiv:2110.08207}, 2021.

\bibitem[Scao et~al.(2022)Scao, Fan, Akiki, Pavlick, Ili{\'c}, Hesslow, Castagn{\'e}, Luccioni, Yvon, Gall{\'e}, et~al.]{scao2022bloom}
Scao, T.~L., Fan, A., Akiki, C., Pavlick, E., Ili{\'c}, S., Hesslow, D., Castagn{\'e}, R., Luccioni, A.~S., Yvon, F., Gall{\'e}, M., et~al.
\newblock Bloom: A 176b-parameter open-access multilingual language model.
\newblock \emph{arXiv preprint arXiv:2211.05100}, 2022.

\bibitem[Sheng et~al.(2023)Sheng, Zheng, Yuan, Li, Ryabinin, Fu, Xie, Chen, Barrett, Gonzalez, et~al.]{sheng2023high}
Sheng, Y., Zheng, L., Yuan, B., Li, Z., Ryabinin, M., Fu, D.~Y., Xie, Z., Chen, B., Barrett, C., Gonzalez, J.~E., et~al.
\newblock High-throughput generative inference of large language models with a single gpu.
\newblock \emph{arXiv preprint arXiv:2303.06865}, 2023.

\bibitem[Singh et~al.(2019)Singh, Natarajan, Shah, Jiang, Chen, Batra, Parikh, and Rohrbach]{singh2019textvqa}
Singh, A., Natarajan, V., Shah, M., Jiang, Y., Chen, X., Batra, D., Parikh, D., and Rohrbach, M.
\newblock Towards vqa models that can read.
\newblock In \emph{Proceedings of the IEEE/CVF conference on computer vision and pattern recognition}, pp.\  8317--8326, 2019.

\bibitem[Taori et~al.(2023)Taori, Gulrajani, Zhang, Dubois, Li, Guestrin, Liang, and Hashimoto]{alpaca}
Taori, R., Gulrajani, I., Zhang, T., Dubois, Y., Li, X., Guestrin, C., Liang, P., and Hashimoto, T.~B.
\newblock Stanford alpaca: An instruction-following llama model.
\newblock \url{https://github.com/tatsu-lab/stanford_alpaca}, 2023.

\bibitem[Tillet et~al.(2019)Tillet, Kung, and Cox]{tillet2019triton}
Tillet, P., Kung, H.-T., and Cox, D.
\newblock Triton: an intermediate language and compiler for tiled neural network computations.
\newblock In \emph{Proceedings of the 3rd ACM SIGPLAN International Workshop on Machine Learning and Programming Languages}, pp.\  10--19, 2019.

\bibitem[Touvron et~al.(2023{\natexlab{a}})Touvron, Lavril, Izacard, Martinet, Lachaux, Lacroix, Rozi{\`e}re, Goyal, Hambro, Azhar, et~al.]{touvron2023llama}
Touvron, H., Lavril, T., Izacard, G., Martinet, X., Lachaux, M.-A., Lacroix, T., Rozi{\`e}re, B., Goyal, N., Hambro, E., Azhar, F., et~al.
\newblock Llama: Open and efficient foundation language models.
\newblock \emph{arXiv preprint arXiv:2302.13971}, 2023{\natexlab{a}}.

\bibitem[Touvron et~al.(2023{\natexlab{b}})Touvron, Martin, Stone, Albert, Almahairi, Babaei, Bashlykov, Batra, Bhargava, Bhosale, et~al.]{touvron2023llama2}
Touvron, H., Martin, L., Stone, K., Albert, P., Almahairi, A., Babaei, Y., Bashlykov, N., Batra, S., Bhargava, P., Bhosale, S., et~al.
\newblock Llama 2: Open foundation and fine-tuned chat models.
\newblock \emph{arXiv preprint arXiv:2307.09288}, 2023{\natexlab{b}}.

\bibitem[Vaswani et~al.(2017)Vaswani, Shazeer, Parmar, Uszkoreit, Jones, Gomez, Kaiser, and Polosukhin]{vaswani2017attention}
Vaswani, A., Shazeer, N., Parmar, N., Uszkoreit, J., Jones, L., Gomez, A.~N., Kaiser, {\L}., and Polosukhin, I.
\newblock Attention is all you need.
\newblock \emph{Advances in neural information processing systems}, 30, 2017.

\bibitem[Wang et~al.(2020)Wang, Zhang, and Han]{spatten}
Wang, H., Zhang, Z., and Han, S.
\newblock Spatten: Efficient sparse attention architecture with cascade token and head pruning.
\newblock \emph{CoRR}, abs/2012.09852, 2020.
\newblock URL \url{https://arxiv.org/abs/2012.09852}.

\bibitem[Wang et~al.(2019)Wang, Liu, Lin, Lin, and Han]{wang2019haq}
Wang, K., Liu, Z., Lin, Y., Lin, J., and Han, S.
\newblock {HAQ: Hardware-Aware Automated Quantization with Mixed Precision}.
\newblock In \emph{CVPR}, 2019.

\bibitem[Wei et~al.(2021)Wei, Bosma, Zhao, Guu, Yu, Lester, Du, Dai, and Le]{wei2021finetuned}
Wei, J., Bosma, M., Zhao, V.~Y., Guu, K., Yu, A.~W., Lester, B., Du, N., Dai, A.~M., and Le, Q.~V.
\newblock Finetuned language models are zero-shot learners.
\newblock \emph{arXiv preprint arXiv:2109.01652}, 2021.

\bibitem[Wei et~al.(2022{\natexlab{a}})Wei, Zhang, Zhang, Gong, Zhang, Zhang, Yu, and Liu]{outlier_suppression}
Wei, X., Zhang, Y., Zhang, X., Gong, R., Zhang, S., Zhang, Q., Yu, F., and Liu, X.
\newblock Outlier suppression: Pushing the limit of low-bit transformer language models, 2022{\natexlab{a}}.
\newblock URL \url{https://arxiv.org/abs/2209.13325}.

\bibitem[Wei et~al.(2022{\natexlab{b}})Wei, Zhang, Zhang, Gong, Zhang, Zhang, Yu, and Liu]{wei2022outlier}
Wei, X., Zhang, Y., Zhang, X., Gong, R., Zhang, S., Zhang, Q., Yu, F., and Liu, X.
\newblock Outlier suppression: Pushing the limit of low-bit transformer language models.
\newblock \emph{arXiv preprint arXiv:2209.13325}, 2022{\natexlab{b}}.

\bibitem[Wei et~al.(2023)Wei, Zhang, Li, Zhang, Gong, Guo, and Liu]{wei2023outlier}
Wei, X., Zhang, Y., Li, Y., Zhang, X., Gong, R., Guo, J., and Liu, X.
\newblock Outlier suppression+: Accurate quantization of large language models by equivalent and optimal shifting and scaling.
\newblock \emph{arXiv preprint arXiv:2304.09145}, 2023.

\bibitem[Xiao et~al.(2022)Xiao, Lin, Seznec, Demouth, and Han]{xiao2022smoothquant}
Xiao, G., Lin, J., Seznec, M., Demouth, J., and Han, S.
\newblock Smoothquant: Accurate and efficient post-training quantization for large language models.
\newblock \emph{arXiv preprint arXiv:2211.10438}, 2022.

\bibitem[Yao et~al.(2022)Yao, Aminabadi, Zhang, Wu, Li, and He]{zeroquant}
Yao, Z., Aminabadi, R.~Y., Zhang, M., Wu, X., Li, C., and He, Y.
\newblock Zeroquant: Efficient and affordable post-training quantization for large-scale transformers, 2022.
\newblock URL \url{https://arxiv.org/abs/2206.01861}.

\bibitem[Yu et~al.(2023)Yu, Yang, Li, Wang, Lin, Liu, Wang, and Wang]{yu2023mmvet}
Yu, W., Yang, Z., Li, L., Wang, J., Lin, K., Liu, Z., Wang, X., and Wang, L.
\newblock Mm-vet: Evaluating large multimodal models for integrated capabilities.
\newblock \emph{arXiv preprint arXiv:2308.02490}, 2023.

\bibitem[Zhang et~al.(2023)Zhang, Han, Zhou, Hu, Yan, Lu, Li, Gao, and Qiao]{zhang2023llama}
Zhang, R., Han, J., Zhou, A., Hu, X., Yan, S., Lu, P., Li, H., Gao, P., and Qiao, Y.
\newblock Llama-adapter: Efficient fine-tuning of language models with zero-init attention.
\newblock \emph{arXiv preprint arXiv:2303.16199}, 2023.

\bibitem[Zhang et~al.(2022)Zhang, Roller, Goyal, Artetxe, Chen, Chen, Dewan, Diab, Li, Lin, Mihaylov, Ott, Shleifer, Shuster, Simig, Koura, Sridhar, Wang, and Zettlemoyer]{opt}
Zhang, S., Roller, S., Goyal, N., Artetxe, M., Chen, M., Chen, S., Dewan, C., Diab, M., Li, X., Lin, X.~V., Mihaylov, T., Ott, M., Shleifer, S., Shuster, K., Simig, D., Koura, P.~S., Sridhar, A., Wang, T., and Zettlemoyer, L.
\newblock Opt: Open pre-trained transformer language models, 2022.
\newblock URL \url{https://arxiv.org/abs/2205.01068}.

\end{thebibliography}
\bibliographystyle{mlsys2024}

\end{document}


\twocolumn[
\mlsystitle{AWQ: Activation-aware Weight Quantization for On-Device LLM Compression and Acceleration}

\begin{mlsysauthorlist}
\mlsysauthor{}{}
\end{mlsysauthorlist}

\mlsyskeywords{Machine Learning, MLSys}

\vskip 0.3in
]
\appendix
\input{text/appendix}
\bibliography{main}
\bibliographystyle{mlsys2024}